%% file: neurips_2026.tex
\documentclass{article}

\PassOptionsToPackage{numbers, square}{natbib}
\usepackage[preprint]{neurips_2026}
\usepackage{wrapfig}
\usepackage{graphicx}
\usepackage{subcaption}
\usepackage{multirow}
\usepackage{tabularx}
\usepackage{titletoc}
\usepackage{longtable}
\usepackage{amsmath}
\usepackage{amssymb}
\usepackage{hyperref}       
\usepackage{mathtools}
\usepackage{amsthm}
\usepackage[capitalize,noabbrev]{cleveref}
\usepackage[textsize=tiny]{todonotes}
\usepackage[utf8]{inputenc} 
\usepackage[T1]{fontenc}    
\usepackage{url}            
\usepackage{booktabs}       
\usepackage{amsfonts}       
\usepackage{nicefrac}       
\usepackage{microtype}      
\usepackage{xcolor}         
\usepackage[table]{xcolor}
\usepackage{colortbl}

\title{Escaping the BLEU Trap: A Signal-Grounded Framework with Decoupled Semantic Guidance for EEG-to-Text Decoding}

\author{
Yuchen Wang$^{1,2}$ \and
Haonan Wang$^{1}$ \and
Yu Guo$^{1}$ \and
Honglong Yang$^{1,2}$ \and
Xiaomeng Li$^{1,2}$\thanks{Corresponding author.}
\\
$^{1}$The Hong Kong University of Science and Technology \\
$^{2}$Shenzhen Loop Area Institute \\
\texttt{ywangzi@connect.ust.hk, eexmli@ust.hk}
}

\begin{document}

\maketitle

\newcommand{\methodname}{\textsc{SemKey}}

\input{sec/0_abstract}   
\input{sec/1_intro}

\input{sec/4_relate_work}
\input{sec/2_method}
\input{sec/3_result}

\input{sec/5_conclusion}

{\small
\bibliographystyle{plainnat}
\bibliography{main}
}


\input{sec/x_appendix}

\newpage
\input{checklist.tex}

\end{document}

%% file: sec/0_abstract.tex
\begin{abstract}
  Decoding natural language from non-invasive EEG signals is a promising yet challenging task. However, current state-of-the-art models remain constrained by three fundamental issues: \textit{Semantic Bias}, where outputs collapse into generic linguistic templates; \textit{Signal Neglect}, where models rely heavily on LLM priors to hallucinate fluent text even in the absence of meaningful signals; and the \textit{``BLEU Trap’’}, where high-frequency stopwords inflate n-gram metrics, masking a lack of true semantic fidelity.
  To resolve these challenges, we move beyond conventional end-to-end pipelines and propose \textbf{\methodname{}}, a novel multi-stage framework that enforces signal-grounded generation through four decoupled semantic objectives: sentiment, topic, length, and surprisal.
  We extract these semantic anchors from EEG embeddings directly, then unify them with an \textit{Active Retrieval Decoding} mechanism, compelling the LLM to ground its token generation in the neural signals rather than defaulting to linguistic priors. Furthermore, we break the BLEU Trap by establishing a comprehensive evaluation protocol using rigorous retrieval and distribution-based metrics such as Fréchet Distance. Extensive experiments demonstrate that \methodname{} effectively mitigates hallucinations on noise inputs and achieves SOTA performance on these robust protocols. Code will be released upon acceptance at \url{https://github.com/xmed-lab/SemKey}.
\end{abstract}

%% file: sec/1_intro.tex
\section{Introduction}
\label{sec:intro}

\begin{figure}[t]
    \centering
    \includegraphics[width=1\linewidth]{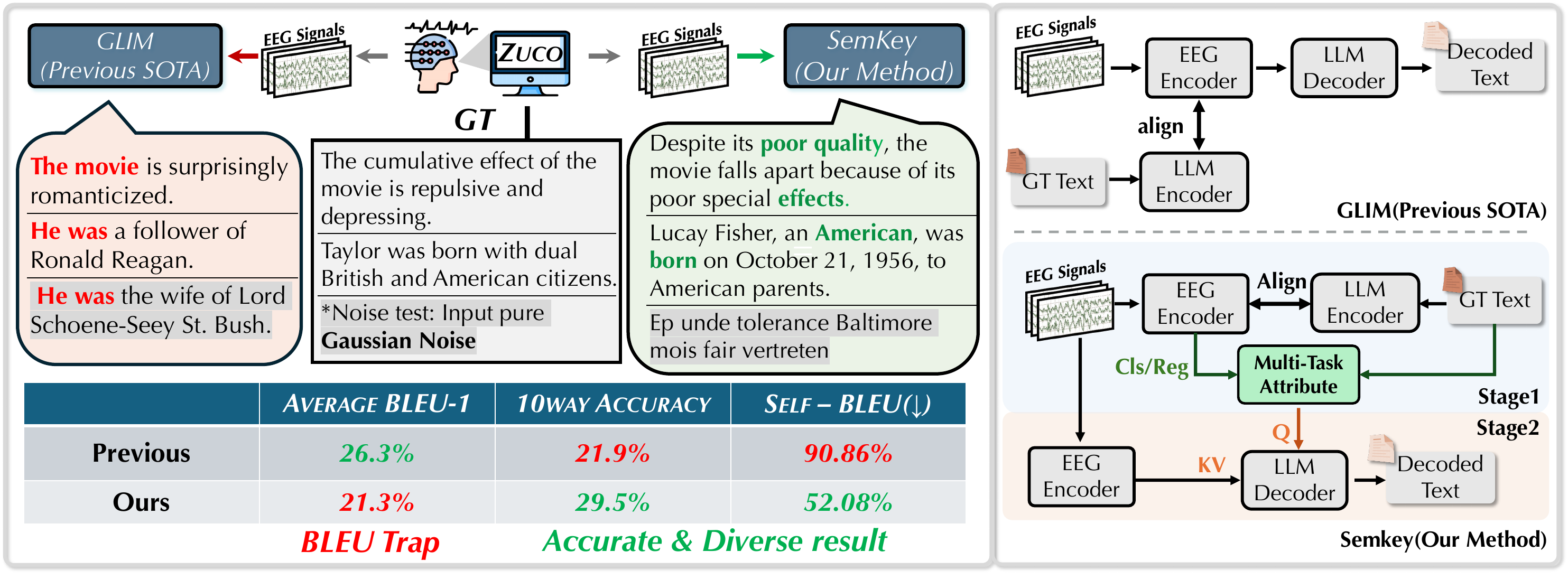}
    \caption{\textbf{Left (Decoding showcase \& Quantitative Results):} Previous models~\cite{liu2025learning} exhibit severe \textit{semantic bias}, overfitting to generic templates to artificially inflate BLEU scores despite low semantic alignment and repetition - a phenomenon we identify as the \textit{``BLEU Trap''}. The noise test (gray highlight) exposes \textit{\textbf{signal neglect}}: the model continues to hallucinate fluent text from pure Gaussian noise. In contrast, \methodname{} breaks the BLEU Trap, demonstrating accurate, diverse, and strictly signal-dependent generation, producing disordered tokens for noise. 
    \textbf{Right (Architectural Comparison):} To address these issues, \methodname{} replaces the conventional end-to-end decoding pipeline (top) with a novel two-stage framework (bottom). }
    \label{fig:Intro}
    \vskip -0.13in
\end{figure}    

Decoding natural language from neural activity is a pivotal frontier in Brain-Computer Interface (BCI) research, offering a lifeline for individuals with aphasia or locked-in syndrome to restore communication \cite{defossez2023decoding, willett2023high}. 
While invasive modalities like Electrocorticography (ECoG) and Stereoelectroencephalography (sEEG) demonstrate high fidelity~\cite{metzger2023high, feng2025acoustic}, their invasive procedures make them unsuitable for widespread use in general-purpose BCI applications.
Consequently, non-invasive Electroencephalography (EEG) remains the most promising modality for widespread adoption due to its portability and safety.

While EEG-to-text decoding holds great promise, the field currently faces a validity crisis. Recent studies~\cite{jo2024eeg} reveal that prior methods~\cite{kostas2021bendr,wang2022open,duan2023dewave,liu2024eeg2text,tao2025see} heavily rely on \textit{Teacher Forcing} (TF) during inference---feeding ground-truth text prefixes to predict subsequent words. This paradigm is fundamentally impractical for real-world Brain-Computer Interfaces (BCIs), where the user's intended message is inherently unknown. Crucially, TF creates a severe learning shortcut: models exploit strong language priors to guess the next word (e.g., predicting ``Washington'' given ``The US president'') rather than actually interpreting the EEG signals. Alarmingly, this textual reliance is so extreme that replacing the training EEG with random noise yields identical performance~\cite{jo2024eeg}. Furthermore, when TF is removed during inference, the generation quality of these prior models degrades rapidly. To ensure outputs are genuinely grounded in neural activity rather than hallucinated from language priors, a deployable BCI must operate without TF. Consequently, our approach strictly enforces fully autoregressive decoding, predicting text based \textit{solely} on input EEG and its own generated tokens.


GLIM~\cite{liu2025learning} is an open-vocabulary method that removes teacher forcing and claims to address the validity crisis in EEG-to-text decoding; however, our analysis reveals that it achieved limited results due to two critical issues: \textbf{semantic bias and signal neglect}. First, GLIM exhibits a severe semantic bias. Its output frequently collapses into highly repetitive, generic linguistic templates (e.g., sentences start with ``He was...'' or ``The movie...'', shown in Figure~\ref{fig:Intro} left). This results in a generated data distribution that diverges significantly from the ground truth (Figure~\ref{fig:tsne} and Appendix Table~\ref{tab:bleu_trap_qualitative}\&\ref{tab:full_qualitative_comparison}), indicating a failure to capture the accurate semantic content from the neural inputs. Second, motivated by recent validity investigations~\cite{jo2024eeg}, our noise input experiments expose that GLIM still heavily relies on linguistic priors from LLM rather than true neural decoding. Specifically, when fed with pure Gaussian noise instead of real EEG signals, the model continues to generate fluent sentences that maintain the identical templated structures seen with real inputs (Figure~\ref{fig:noise_case} and Table~\ref{tab:noise_samples}).

Motivated by these observations, we propose a novel method \methodname{}, which features a \textbf{Parallel Neural-Driven Attribute Decouple} module to enhance the model's semantic capacity in decoding EEG. This module is motivated by neuroscientific research demonstrating that high-level semantics form robust cortical representations~\cite{lerner2011topographic}. To mitigate the aforementioned semantic bias, we explicitly disentangle key semantic attributes (\textit{sentiment, topic, length, surprisal}) directly from the EEG features (Figure~\ref{fig:Intro} right bottom). By capturing these high-level semantic context, we establish strong generative boundaries that effectively constrain the LLM and prevent template collapse. In addition, to address signal neglect, we implement an \textbf{Active Retrieval Decoding} mechanism. By adjusting the Q-K-V workflow, this mechanism actively fuses the disentangled semantic attributes and continuous EEG features into the decoding process, compelling the LLM to ground its token generation directly in the provided neural signals.

Furthermore, we observe that the template memorization behavior of existing baselines is often obscured by the \textit{\textbf{``BLEU Trap''}}(Figure~\ref{fig:Intro}). Conventional evaluation protocols that rely heavily on n-gram matching metrics are fundamentally limited. Models can achieve competitive scores by simply replicating stopwords and frequent templates, artificially inflating performance while masking a lack of semantic fidelity (Table~\ref{tab:bleu_trap_qualitative}). To resolve this, we move beyond the BLEU Trap by establishing a comprehensive evaluation protocol utilizing rigorous retrieval-based and distribution-based metrics. Comprehensive evaluations demonstrate that \methodname{} establishes a new state-of-the-art, prominently outperforming the GLIM baseline.

%% file: sec/4_relate_work.tex
\section{Related Work}
\label{sec:related_work}

\textbf{EEG-to-Text Decoding: Evolution and Frontiers.} 
Early approaches relied on handcrafted features like Mel-frequency cepstral coefficients (MFCCs)~\cite{cooney2018mel} and wavelet transforms~\cite{panachakel2020novel}, necessitating extensive domain expertise. The advent of deep learning introduced automated feature extraction via CNNs and LSTMs~\cite{zhang2018converting} to capture spatio-temporal dynamics. A paradigm shift occurred with Transformer-based pipelines: \cite{wang2022open} proposed projecting EEG signals into a pre-trained BART latent space, and building on this, DeWave~\cite{duan2023dewave} introduced an end-to-end framework processing raw brainwaves for direct translation.
More recently, the integration of Large Language Models has spurred new directions. Thought2Text~\cite{mishra2025thought2text} explores leveraging LLMs for EEG encoding, while WaveMind~\cite{zeng2025wavemind} proposes a conversational foundation model aligned across textual and visual modalities. \textit{However, our work differs from these distinct streams:} unlike WaveMind and Thought2Text, which relies on \textbf{image stimuli} that offer richer and more direct semantic grounding, we tackle the different task of \textit{pure} EEG-to-Text decoding from abstract \textit{textual stimuli} without such visual aids. 
\textbf{The Validity Crisis and Methodological Gap.} 
Notwithstanding architectural advancements, the field faces a crisis of validity. The pivotal study ``Are EEG-to-Text Models Working''~\cite{jo2024eeg} exposed that high BLEU scores in prior arts are often artifacts of \textit{Teacher Forcing}, where models ignore the noisy EEG encoder in favor of the LLM's linguistic priors. While GLIM~\cite{liu2025learning} recently addressed this by establishing the first effective pipeline without teacher forcing, it—along with its predecessors—still suffers from ``signal neglect,'' leading to semantic collapse in open-ended generation (as discussed in Section~\ref{sec:intro}).
To bridge this gap between open-ended generation and faithful neural decoding, \methodname{} departs from the prevailing reliance on implicit latent alignment. We propose a \textit{Coarse-to-Fine} decoding paradigm to enforce validity: First, we introduce an \textit{explicit, attribute-driven} mechanism where high-level semantic anchors (e.g., Sentiment, Topic) actively steer the decoding trajectory. This serves as a hard constraint to prevent semantic collapse, ensuring adherence to the global semantic intent. Complementing this high-level guidance, we adjust the fine-grained LLM interaction via a \textit{Q-K-V Injection} workflow. By forcing EEG embeddings to act as Keys and Values, we enforce a strict token-level dependency on neural signals.

%% file: sec/2_method.tex
\section{Method}

\methodname{} is a two-stage framework designed to mitigate semantic bias and signal neglect.
It comprises: (1) a \textit{Parallel Neural-Driven Attribute Decouple} stage that learns robust EEG representations aligned with the linguistic latent space while explicitly decoupling high-level semantic attributes; and (2) a \textit{Multi-Perspective Active Retrieval Decoding} stage that actively exploits these representations to produce coherent and faithful text.

 \begin{figure*}[ht]
    \centering
    \includegraphics[width=1\linewidth]{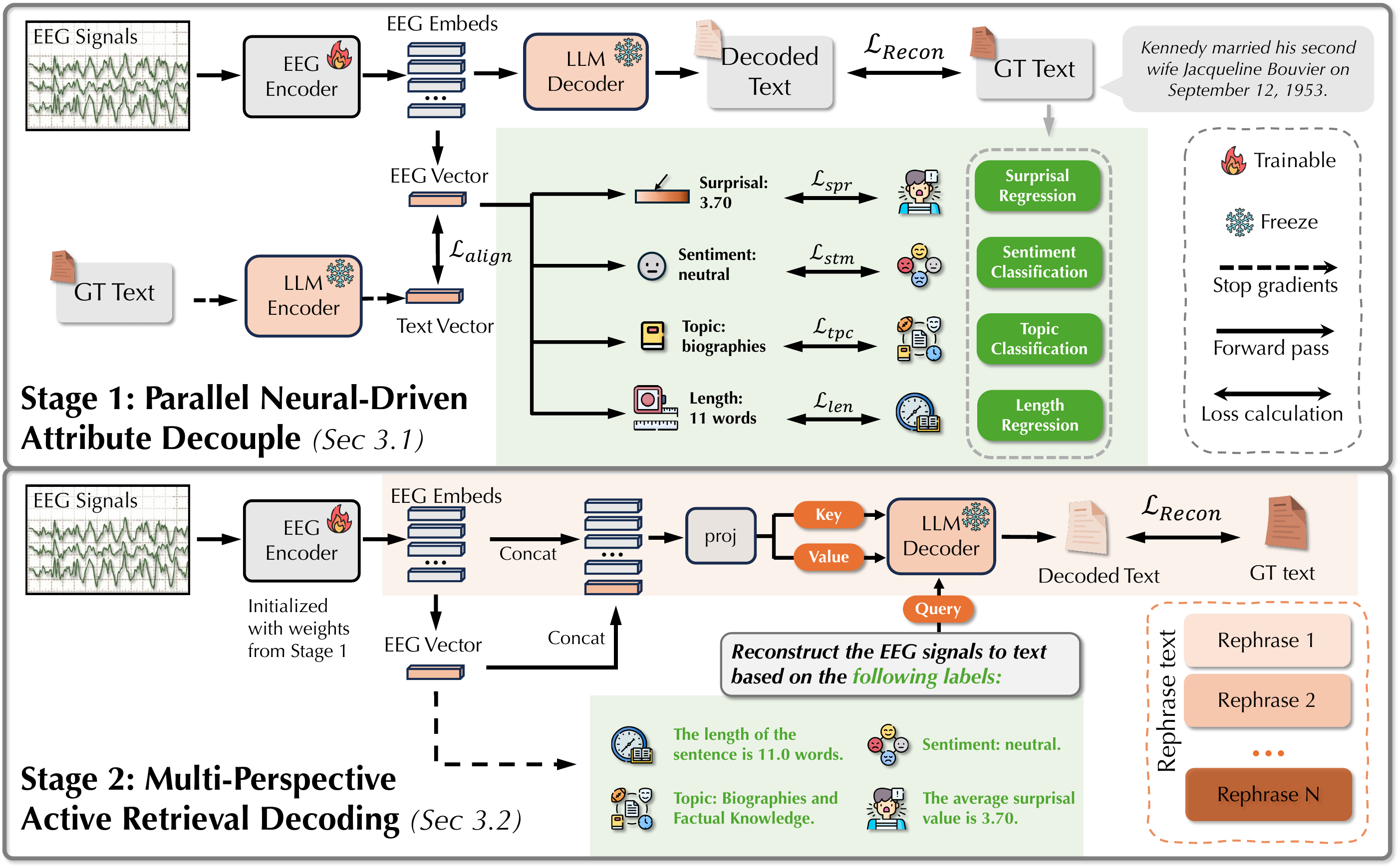}
    \caption{The overall architecture of the \methodname{} framework following a "Guidance-Generation" paradigm. \textbf{Stage 1 (Attribute Extraction)}: The EEG encoder is optimized via multi-task learning to explicitly decouple high-level semantic attributes (Sentiment, Topic, Length, Surprisal) alongside standard alignment. \textbf{Stage 2 (Generative Decoding)}: These predicted attributes structure a semantic prompt to guide generation. Crucially, the \textbf{Q-K-V Injection} workflow enforces signal dependency by using the text prompt as the Query and projected EEG embeddings as Keys and Values.}
    \label{fig:structure}
\end{figure*}

\subsection{Stage 1: Parallel Neural-Driven Attribute Decouple}

Based on \cite{liu2025learning}, we adopt a multi-task learning framework to explicitly disentangle high-level semantics from EEG representations. The EEG encoder maps signals into latent embeddings, inheriting alignment and reconstruction objectives via a frozen Flan-T5 model \cite{chung2024flant5}. In addition, four auxiliary heads predict sentiment, topic, length, and surprisal, forming a structured semantic anchor for downstream generation.

\noindent\textbf{Prior-Guided EEG Encoding Stream.}
Following GLIM \cite{liu2025learning}, we mitigate inter-subject variability by embedding discrete priors (e.g., subject ID, dataset) into a context vector. This vector modulates a Conformer encoder \cite{gulati2020conformer, jiang2024labram} that processes input signals to output two representations: (1) \textbf{EEG Embeddings} $\mathbf{E}_{eeg} \in \mathbb{R}^{L \times d_{model}}$ for token-level reconstruction, and (2) an attention-pooled \cite{lee2019set} \textbf{EEG Vector} $\mathbf{v}_{eeg} \in \mathbb{R}^{d_{model}}$ for high-level semantic prediction. The encoder is optimized using reconstruction ($\mathcal{L}_{recon}$) and alignment ($\mathcal{L}_{align}$) losses to ensure text-space consistency \cite{liu2025learning}.

\noindent\textbf{Parallel Neural-Driven Learning Framework.}
To enrich the semantic structure of $\mathbf{v}_{eeg}$, we introduce a parallel multi-task framework \cite{wang2025neurons}. Four attribute heads are attached to $\mathbf{v}_{eeg}$. Crucially, these four attributes constitute an complementary semantic anchor that comprehensively circumscribes the generation process. \textit{Topic} and \textit{Sentiment} act as macroscopic anchors that dictate the semantic category and affective tone, directly preventing the LLM from collapsing into irrelevant generic templates (e.g., semantic bias)~\cite{liang2024controllable,lorandi2023control}. \textit{Length} serves as a structural boundary, reflecting the working memory load during language processing~\cite{just1992capacity}, which effectively curtails recursive loops and repetitive generation. Furthermore, \textit{Surprisal} (information density)—which correlates strongly with well-documented ERP components like the N400~\cite{kutas1980reading}—forces the LLM to modulate its probability distribution~\cite{michaelov2024strong}. This explicitly guides the model to generate diverse, contextually appropriate words rather than defaulting to high-frequency ``safe'' tokens, thereby breaking the BLEU Trap. We formulate the specific objectives as follows:

\textit{Classification Objectives (Sentiment \& Topic):}
Two classifiers predict discrete labels, optimized using cross-entropy loss
$\mathcal{L}_{stm/tpc} = - \sum_{c=1}^{N_C} y_c \log(\hat{y}_c)$.
The Topic labels are [Biographies and Factual Knowledge, Movie Reviews and Sentiment], following the official ZuCo corpus categorization, while the Sentiment labels are [non\_neutral, neutral].

\textit{Regression Objectives (Length \& Surprisal):}
Two regressors predict continuous attributes, optimized using mean squared error
$\mathcal{L}_{len/spr} = \frac{1}{N} \sum_{i=1}^{N} (y_i - \hat{y}_i)^2$,
with normalized targets for stable training.

\noindent\textbf{Overall Training Objective.}
The overall objective is:

\begin{equation}
\mathcal{L}_{Stage1} = \lambda_{align}\mathcal{L}_{align} + \lambda_{recon}\mathcal{L}_{recon} 
+ \lambda_{cls}(\mathcal{L}_{stm} + \mathcal{L}_{tpc}) + \lambda_{reg}(\mathcal{L}_{len} + \mathcal{L}_{spr})
\end{equation}

where $\lambda$ are hyperparameters balancing each component (detailed in Appendix~\ref{app:implementation_details}).This composite objective results in an EEG Encoder capable of producing semantically disentangled and attribute-aware embeddings, laying the foundation for the generative stage.

\subsection{Stage 2: Multi-Perspective Active Retrieval Decoding}
Prediction heads from Stage 1 are frozen, and the aligned EEG features are injected into the LLM via a learnable projector. To ensure fidelity, we introduce two key mechanisms: (1) \textit{Prompt Guidance}, where predicted attributes are structured into a natural language constraint; and (2) \textit{Q-K-V Injection}, which re-routes the attention workflow: the text prompt serves as the Query, interacting with EEG embeddings and EEG vector as Keys and Values. This transforms passive conditioning into an \textit{active retrieval} process, compelling the LLM to actively query neural signals for every token. The EEG Encoder remains trainable while the LLM Decoder is frozen.

\noindent\textbf{Multi-Perspective Prompt.}
To explicitly guide text reconstruction using the high-level semantics obtained in Stage 1, we leverage the decoupled attribute predictions to construct a structured natural language prompt, serving as a ``semantic anchor'' for the LLM \cite{keskar2019ctrl, lester2021power}.Let $\mathcal{H} = \{\hat{y}_{len}, \hat{y}_{tpc}, \hat{y}_{stm}, \hat{y}_{spr}\}$ be the set of predicted attributes. A template function $\mathcal{T}(\cdot)$ maps these attributes into a prompt $\mathbf{X}_{prompt} = \mathcal{T}(\mathcal{H})$, the specific prompts can be found in Appendix~\ref{app:implementation_details}.
This prompt $\mathbf{X}_{prompt}$ is then tokenized and prepended to the decoder input. The hidden states derived from this prompt act as the \textbf{Queries ($Q$)}, seeking relevant information to fulfill semantic constraints.

\noindent\textbf{Q-K-V Injection Workflow.}
Unlike previous approaches that concatenate brain embeddings with text tokens—leading to signal neglect via self-attention \cite{wu2022characterizing, mishra2025thought2text}—we employ a \textit{Q-K-V Injection} strategy to separate the information flow \cite{li2023blip, alayrac2022flamingo}. The autoregressive input sequence consists of the prompt $\mathbf{X}_{prompt}$ and previously generated tokens $\mathbf{y}_{<t}$. The decoder processes this to obtain contextualized hidden states $\mathbf{H}_{text} = \Phi_{self}([\mathbf{X}_{prompt}; \mathbf{y}_{<t}])$, which serve as \textit{Queries} for the cross-attention layers. To provide a multi-scale representation, we prepend the global vector $\mathbf{v}_{eeg}$ to the token-level embeddings $\mathbf{E}_{eeg}$, forming an augmented $(L+1) \times d_{model}$ sequence. This unified representation is then projected to form the \textit{Keys} and \textit{Values}:
\begin{equation}
    Q = \mathbf{H}_{text} \mathbf{W}_Q, \quad K = ([\mathbf{v}_{eeg}; \mathbf{E}_{eeg}]\mathbf{W}_{proj}) \mathbf{W}_K, \quad V = ([\mathbf{v}_{eeg}; \mathbf{E}_{eeg}]\mathbf{W}_{proj}) \mathbf{W}_V
\end{equation}


\subsection{Escaping the ``BLEU Trap'': A Holistic Evaluation Protocol}
\label{sec:metrics}

\noindent\textbf{Retrieval-based Alignment Metrics ($N$-Way Accuracy).}
To measure semantic alignment, we employ \textit{$N$-way Retrieval Accuracy} \cite{radford2021learning, li2024visual}. We utilize the \texttt{sentence-transformers} \cite{reimers2019sentence} library to encode the generated and ground-truth texts into dense vectors.

To mitigate variance from random negative sampling, we adopt a Monte Carlo sampling approach, conducting $M=1000$ independent trials for each difficulty level ($N \in \{2, 4, 10, 24\}$). In each trial $m$, we evaluate one generated text embedding $\mathbf{v}_{gen}^{(m)}$ against its corresponding ground-truth embedding $\mathbf{v}_{gt}^{(m)}$ and a distractor set $\mathcal{N}_{neg}^{(m)}$ containing $N-1$ randomly sampled negative vectors. The final accuracy is averaged over $M$ trials:
\begin{equation}
    \text{Acc}_N = \frac{1}{M} \sum_{m=1}^{M} \mathbb{I} \Big( \text{CosSim}(\mathbf{v}_{gen}^{(m)}, \mathbf{v}_{gt}^{(m)}) > \max_{\mathbf{v}_{k} \in \mathcal{N}_{neg}^{(m)}} \text{CosSim}(\mathbf{v}_{gen}^{(m)}, \mathbf{v}_{k}) \Big)
\end{equation}
where $\mathbb{I}(\cdot)$ is the indicator function. This metric strictly evaluates the model's ability to discriminate the correct semantic content from candidate distractors.

\noindent\textbf{Semantic Substance Metric (Content Recall).}
High BLEU scores can be achieved by matching stopwords, while missing the core meaning. To evaluate the recovery of important information, we introduce \textit{Content Recall}.
We first filter out stopwords from both the reference and generated text, retaining only content-bearing words (nouns, verbs, adjectives). Content Recall calculates the percentage of ground-truth content words that successfully appear in the generated hypothesis. This metric penalizes ``fluent but empty'' hallucinations and rewards semantic fidelity \cite{ji2023survey, maynez2020faithfulness}.

\noindent\textbf{Generative Diversity \& Quality Metrics.}
A major failure mode of previous EEG-to-Text models is \textit{semantic bias}, where the model repeats a few safe sentences. To explicitly break this pattern of repetition and template outputs, we quantify generative diversity and quality using three key metrics. 

First, to evaluate generative diversity, we employ \textit{Self-BLEU} \cite{zhu2018texygen} and \textit{Head Entropy (Prefix Diversity)} \cite{holtzman2019curious}. Self-BLEU measures the BLEU score of each generated sentence against all others in the test set, while Head Entropy calculates the Shannon entropy of the initial bigrams (first two tokens) across all generated sentences. Crucially, these two metrics are not designed to pursue absolute extreme values (e.g., arbitrarily low Self-BLEU or infinitely high entropy). Instead, the goal is for the model's scores to closely \textit{match the values of the ground-truth} (GT) text. Approximating the GT diversity ensures that the model generates uniquely structured sentences for different EEG inputs—effectively avoiding regurgitated memorized templates—while maintaining natural linguistic coherence.

Finally, to evaluate semantic realism at the distribution level, we compute the \textit{Fréchet Distance (FD)} \cite{semeniuta2018fd}. We map all sentences to a high-dimensional semantic space using a pre-trained Sentence Transformer and calculate the distance between the multivariate Gaussian distributions of the generated ($\mu_g, \Sigma_g$) and reference ($\mu_r, \Sigma_r$) embeddings:
\begin{equation}
    \text{FD} = ||\mu_r - \mu_g||^2 + \text{Tr}(\Sigma_r + \Sigma_g - 2(\Sigma_r \Sigma_g)^{1/2})
\end{equation}
A lower FD indicates that the distribution of the decoded text closely matches the semantic manifold of the ground truth text.

%% file: sec/3_result.tex
\section{Experiments \& Results}
\label{sec:experiments}

\begin{wrapfigure}{r}{0.6\textwidth}
    \centering
    \includegraphics[width=\linewidth]{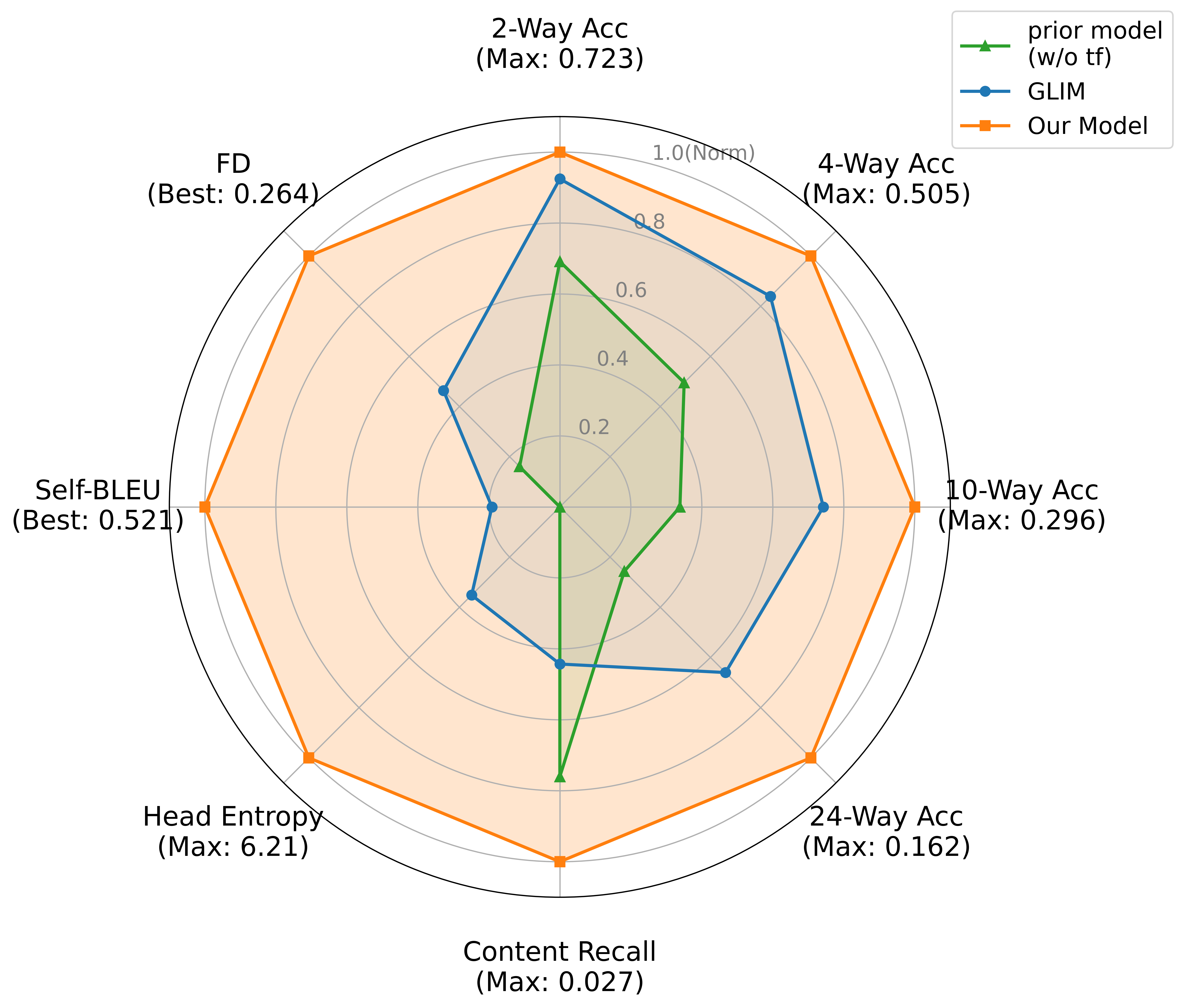}
    \caption{\textbf{Holistic Performance Evaluation.} The Radar Chart illustrates the relative performance of our model (Orange) versus GLIM (Blue) and prior model without teacher forcing (Green). \methodname{} demonstrates a balanced superiority across all alignment, diversity, and content quality metrics.}
    \label{fig:radar}
\end{wrapfigure}

We conduct our experiments using the data from the \textit{ZuCo 1.0}~\cite{hollenstein2018zuco} and \textit{ZuCo 2.0}~\cite{hollenstein2020zuco} datasets, containing over 22K sentence-EEG pairs (detailed in Appendix~\ref{app:dataset}). This dataset is widely adopted by previous state-of-the-art methods to validate the models' cross-subject and cross-dataset generalization capabilities\cite{duan2023dewave,liu2024eeg2text}. For comprehensive details regarding dataset splits, preprocessing protocols, and implementation specifics (including optimizer settings and hyperparameters), please refer to \textit{Appendix~\ref{app:implementation_details}}.

We compare \textit{\methodname{}} against the recent baseline \textit{GLIM}~\cite{liu2025learning}, along with previous methods including \textit{EEG-to-Text}~\cite{wang2022open}, \textit{DeWave}~\cite{duan2023dewave}, and \textit{EEG2Text}~\cite{liu2024eeg2text}. As discussed in~\cite{jo2024eeg}, evaluating these prior models with the teacher forcing (TF) mechanism introduces information leakage, making them inapplicable to real-world scenarios where ground-truth prefixes are unavailable. Because TF-based evaluations are not our primary experimental focus, we only report the full comprehensive metrics for \textit{EEG-to-Text} under this setting. When teacher forcing is removed to simulate practical auto-regressive inference, these prior models suffer from full signal neglect and exhibit same performance~\cite{jo2024eeg}. Therefore, we consolidate them into a single representative baseline, \textit{Prior Model (w/o TF)}, which is instantiated solely by \textit{EEG-to-Text} since it is the only fully open-source model among the three. Furthermore, to measure the upper bound of our proposed evaluation protocol, we also report the performance using ground-truth (GT) inputs. All experiments were conducted using different random seeds, and we performed hypothesis testing to ensure that our improvements are statistically significant (detailed results in Appendix~\ref{app:seed}).

\subsection{Main Performance Comparison}
\label{sec:main_results}

\begin{table*}[ht]
\caption{Performance comparison on BLEU and new proposed metric. All experiments are conducted on the ZuCo datasets. We prioritize the results reported in the original papers; for our newly proposed metrics, we utilize the officially released repositories for reproduction and evaluation. Best results are in bold. ``Prior Model (w/o TF)'' represents the previous methods evaluated without teacher-forcing to reflect their true generation capability. The ``GT'' row serves as an upper bound, where the ground truth is directly used as the input. Gray rows indicate different evaluation settings, while gray text represents traditional BLEU metrics.}
\label{tab:main_results}
\begin{center}
\begin{sc}
\makebox[\textwidth][c]{
\resizebox{\textwidth}{!}{
\renewcommand{\arraystretch}{1.35}
\setlength{\tabcolsep}{2pt} 
\begin{tabular}{l cccc cccccccc} 
\toprule
& \multicolumn{4}{c}{BLEU Metrics (\%)} & \multicolumn{4}{c}{Retrieval Accuracy (\%)} & Substance & Diversity & \multicolumn{2}{c}{Quality ($\downarrow$)} \\
\cmidrule(lr){2-5} \cmidrule(lr){6-9} \cmidrule(lr){12-13}
Model & B-1 & B-2 & B-3 & B-4 & 2-Way & 4-Way & 10-Way & 24-Way & C. Rec. (\%) & H. Ent & S-BLEU (\%) & FD \\
\midrule
\rowcolor{gray!20} \multicolumn{13}{c}{\textit{With Teacher Forcing}} \\
EEG-to-Text~\cite{wang2022open} & \textcolor{gray}{40.12} & \textcolor{gray}{23.18} & \textcolor{gray}{12.61} & \textcolor{gray}{6.80} & 95.66 & 90.06 & 80.30 & 68.48 & 16.92 & 6.06 & 78.39 & 0.76 \\
DeWave~\cite{duan2023dewave} & \textcolor{gray}{41.35} & \textcolor{gray}{24.15} & \textcolor{gray}{13.92} & \textcolor{gray}{8.22} & - & - & - & - & - & - & - & - \\
EEG2Text~\cite{liu2024eeg2text} & \textcolor{gray}{45.20} & \textcolor{gray}{29.10} & \textcolor{gray}{19.70} & \textcolor{gray}{14.10} & - & - & - & - & - & - & - & - \\
\midrule
\rowcolor{gray!20} \multicolumn{13}{c}{\textit{Without Teacher Forcing}} \\
\begin{tabular}[c]{@{}l@{}}Prior Model \\ (w/o TF)\cite{wang2022open,duan2023dewave,liu2024eeg2text}\end{tabular} & \textcolor{gray}{22.01} & \textcolor{gray}{6.17} & \textcolor{gray}{2.56} & \textcolor{gray}{1.26} & 50.00 & 25.00 & 10.00 & 4.17 & 2.04 & 0.00 & 100 & 1.65 \\
GLIM~\cite{liu2025learning} & \textcolor{gray}{26.32} & \textcolor{gray}{10.67} & \textcolor{gray}{3.96} & \textcolor{gray}{1.86} & 66.83 & 42.45 & 21.91 & 10.74 & 1.24 & 2.18 & 90.86 & 0.57 \\
\methodname{}(Ours) & \textcolor{gray}{21.30} & \textcolor{gray}{6.39} & \textcolor{gray}{1.82} & \textcolor{gray}{0.81} & \textbf{72.29} & \textbf{50.48} & \textbf{29.56} & \textbf{16.24} & \textbf{2.71} & \textbf{6.21} & \textbf{52.08} & \textbf{0.26} \\
\midrule
GT (Upper Bound) & \textcolor{gray}{100} & \textcolor{gray}{100} & \textcolor{gray}{100} & \textcolor{gray}{100} & 99.49 & 98.62 & 95.88 & 89.85 & 100 & 6.35 & 49.38 & 0.00 \\
\bottomrule
\end{tabular}
}
}
\end{sc}
\end{center}
\end{table*}

\textbf{Quantitative Results.} Table \ref{tab:main_results} and Figure \ref{fig:radar} present the comprehensive performance comparison. \methodname{} demonstrates robust improvements across three key dimensions.
First, \methodname{} consistently outperforms baselines across all retrieval settings ($N=2, 4, 10, 24$), with the advantage becoming more pronounced as task difficulty increases. This indicates that our model is substantially more robust in distinguishing fine-grained semantic nuances against a large pool of distractors. 
Second, in terms of \textit{Generative Diversity}, \methodname{} exhibits a substantial leap in structural variety (e.g., higher \textit{Head Entropy}). This confirms that our model effectively avoids the ``safe loop''—such as repetitively generating ``He was...''—and produces diverse sentence openings. 
Finally, \methodname{} significantly improves \textit{Generative Quality} across both \textit{Self-BLEU} and \textit{FD}, demonstrating that our generated semantic distribution is substantially closer to the ground truth manifold.

It is worth noting that while the \textit{Prior Model (w/o TF)} suffers from complete signal neglect (outputting identical sentences for all inputs), it paradoxically achieves a higher \textit{Content Recall} than GLIM. This artifact stems from memorizing high-frequency words from the training corpus rather than performing genuine decoding. Its chance-level \textit{Retrieval Accuracy} underscores the necessity of a \textit{holistic evaluation}: high recall alone is insufficient and must be interpreted alongside accuracy and diversity to rule out hallucination.
Similarly, the teacher-forced model (\textit{EEG-to-Text}) exhibits abnormal behavior. Despite high \textit{Retrieval Accuracy} and \textit{Content Recall}, its generation quality and diversity suffer severe degradation, evidenced by a substantially worse \textit{FD} and an inflated \textit{Self-BLEU} score compared to \methodname{}. Manual inspection reveals that this discrepancy is caused by the model generating massive meaningless suffixes (e.g., ``ggg,,,,''). This phenomenon highlights the advantage of our comprehensive evaluation protocol: it can successfully detect underlying mode collapse issues that traditional metrics like BLEU fail to capture.

\begin{table}[ht]
  \vskip -0.15in
  \caption{BLEU metrics comparison (BLEU-$N$). Note that ``MTV Evaluation'' here refers to the use of augmented references during scoring. The significant performance gap observed upon removing MTV indicates that the MTV evaluation setting is a key reason for the BLEU Trap.}
  \label{tab:bleu_scores}
  \begin{center}
    \begin{small}
      \begin{sc}
        \resizebox{0.6\columnwidth}{!}{
        \begin{tabular}{lcccc}
          \toprule
          & \multicolumn{2}{c}{MTV Evaluation} & \multicolumn{2}{c}{Standard (w/o MTV)} \\
          \cmidrule(lr){2-3} \cmidrule(lr){4-5}
          Model & B-1 & B-2 & B-1 & B-2 \\
          \midrule
          \begin{tabular}[c]{@{}l@{}}Prior Model(w/o tf) \\ \cite{wang2022open,duan2023dewave,liu2024eeg2text}\end{tabular} 
            & \underline{22.00\%} & 6.17\% & \textbf{11.90\%} & \textbf{1.58\%} \\
          \begin{tabular}[c]{@{}l@{}}GLIM\cite{liu2025learning}\end{tabular} 
            & \textbf{26.30\%} & \textbf{10.70\%} & 7.84\% & 1.28\% \\
          \midrule
          \methodname{}(Ours) 
            & 21.30\% & \underline{6.39\%} & \underline{8.98\%} & \underline{1.42\%} \\
          \bottomrule
        \end{tabular}
        }
      \end{sc}
    \end{small}
  \end{center}
\end{table}

\noindent\textbf{The ``BLEU Trap'': Metric Inflation via Semantic Bias.}
\label{sec:bleu-trap}
As shown in Table \ref{tab:bleu_scores}, our results empirically confirm the ``BLEU Trap,'' where evaluation metrics are inflated by linguistic priors. A key driver of this trap is the widespread \textit{MTV protocol}, which computes scores against an augmented set of paraphrases. This expanded reference pool disproportionately rewards generic templates, allowing baselines like GLIM to inflate scores by hallucinating ``safe,'' high-frequency prefixes (e.g., ``The movie...'').
Removing these augmented references in the \textit{Standard setting (w/o MTV)} causes a drastic performance shift. GLIM's scores degrade severely, falling behind our proposed \methodname{}. This sharp decline reveals that GLIM's apparent advantage under the MTV protocol is largely an artifact of matching generic patterns rather than performing faithful semantic decoding.

More intriguingly, under this Standard setting, the \textit{Prior Model (w/o TF)} paradoxically achieves the highest BLEU scores. Given that this baseline suffers from complete mode collapse, its top ranking constitutes compelling evidence of the BLEU Trap. It demonstrates that the metric fails to penalize semantic vacuity and can be artificially inflated by repetitively generating high-frequency words, thereby masking the model's fundamental inability to capture unique semantic content. Extended analysis and qualitative examples are provided in Appendix~\ref{app:mtv_bias} and Appendix~\ref{app:bleu_trap}.

\begin{table}[ht]
  \label{tab:noise-in}
  \vspace{-0.1in}
  \caption{Quantitative verification of signal dependency. We compare performance under Real EEG vs. Gaussian Noise. A larger drop in \methodname{} indicates stronger dependency on neural signals, whereas baselines show signs of overfitting to language priors (hallucination).}
  \label{tab:noise_experiment}
  \begin{center}
    \begin{small}
      \begin{sc}
        \resizebox{0.65\columnwidth}{!}{
        \begin{tabular}{llccc}
          \toprule
          Model & Input & C. Recall & Head Entropy & FD \\
          \midrule
          \multirow{2}{*}{\shortstack[l]{Prior Model(w/o tf) \\ \cite{wang2022open,duan2023dewave,liu2024eeg2text}}} 
            & Real  & 2.04\% & 0.00 & 1.65 \\
            & Noise & 2.04\% & 0.00 & 1.65 \\
          \midrule
          \multirow{2}{*}{\shortstack[l]{GLIM \cite{liu2025learning}}} 
            & Real  & 1.24\% & 2.18 & 0.57 \\
            & Noise & 0.56\% & 5.65 & 0.72 \\ 
          \midrule
          \multirow{2}{*}{\methodname{}(Ours)} 
            & Real  & \textbf{2.71\%} & \textbf{6.21} & \textbf{0.26} \\
            & Noise & \textbf{0.04\%} & \textbf{11.08} & \textbf{1.27} \\ 
          \bottomrule
        \end{tabular}
        }
      \end{sc}
    \end{small}
  \end{center}
\end{table}

\subsection{Verification of Signal Dependency}
\label{sec:noise_test}

\begin{wrapfigure}{R}{0.5\textwidth}
\centering
\includegraphics[width=\linewidth]{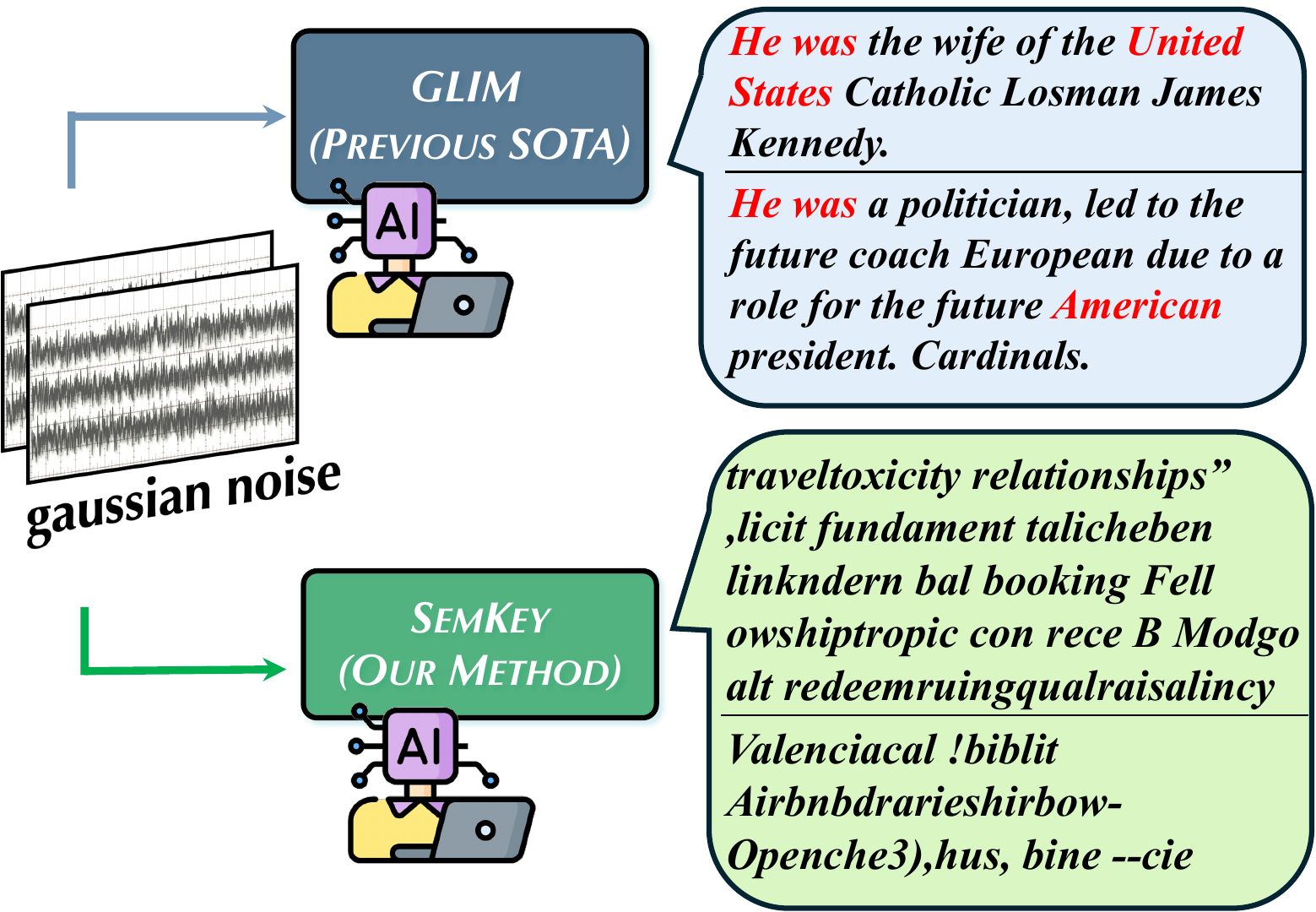}
\caption{\textbf{Showcase of Gaussian Noise input.} We illustrate the text generated by GLIM versus our method when the input is pure Gaussian noise instead of EEG signals. }
\label{fig:noise_case}
\end{wrapfigure}

To verify the signal dependency of our framework, we conducted a \textit{Noise Input Test}, where real EEG signals were replaced with Gaussian noise during inference.

\noindent\textbf{Quantitative Analysis: The ``Garbage In, Garbage Out'' Test.}
We compare the performance drop between Real EEG and Gaussian Noise inputs across three models (Table \ref{tab:noise_experiment}). The results reveal severe signal neglect in baselines: the \textit{Prior Model (w/o TF)} yields identical metrics regardless of whether the input is real brainwaves or pure noise. Similarly, while \textit{GLIM} shows a drop in \textit{Content Recall}, it maintains a relatively low \textit{FD}, indicating a persistent reliance on language priors to hallucinate plausible sentences despite the noisy input. In contrast, \textit{\methodname{}} demonstrates \textit{performance collapse} characteristic of strict signal dependency. When fed with noise, its \textit{Content Recall} vanishes to nearly zero, and its \textit{Head Entropy} and \textit{FD} rises drastically. This extreme entropy reflects \textit{unstructured chaos} rather than linguistic richness, empirically proving that our Q-K-V injection workflow strictly prevents the generation of coherent text in the absence of valid neural information. Additionally, we conducted an experiment involving a gradual signal destruction process to further validate our method's strict signal dependency (see Appendix~\ref{app:noise_test}).

\noindent\textbf{Qualitative Case Study.}
To provide a more intuitive understanding of the decoding behavior, we visualize representative qualitative examples under Gaussian noise input in Figure \ref{fig:noise_case}. As observed, even when fed with pure Gaussian noise, the baseline GLIM continues to generate grammatically fluent but semantically vacuous sentences, such as \textit{``He was ...''}. This phenomenon confirms that the model is over-relying on high-frequency linguistic patterns and language priors rather than decoding the actual neural signals. Conversely, \methodname{} outputs unstructured character sequences and gibberish , correctly reflecting the chaotic nature of the Gaussian noise input. This explicitly validates that our framework enforces strict signal dependency and avoids hallucinating content in the absence of valid neural information. More examples can be found in Appendix~\ref{app:appendix_noise_samples}.

\begin{table}[ht]
\centering
\vskip -0.1in

\begin{minipage}[t]{0.4\columnwidth}
\centering
\caption{Performance of Stage 1 decoupled attribute prediction heads.}
\label{tab:attributes}
\resizebox{\linewidth}{!}{
\begin{tabular}{llcc} 
\toprule
Task & Metric & Chance & Ours \\
\midrule
Topic     & Acc ($\uparrow$) & 50\% & \textbf{83.51\%}  \\
Sentiment & Acc ($\uparrow$) & 50\% & \textbf{90.71\%} \\
Length    & MAE ($\downarrow$) & 6.23  & \textbf{5.01}  \\
Surprisal & MAE ($\downarrow$) & 0.78  & \textbf{0.58}   \\
\bottomrule
\end{tabular}
}
\end{minipage}
\hfill
\begin{minipage}[t]{0.5\columnwidth}
\centering
\caption{Ablation study on key components.}
\label{tab:ablation_full}
\resizebox{\linewidth}{!}{
\begin{tabular}{lcccc}
\toprule
Variant & 4-Way & 24-Way & Self-BLEU($\downarrow$) & FD($\downarrow$) \\
\midrule
w/o Q-K-V & 39.8\% & 9.2\% & 80.9\% & 0.66 \\
w/o Prompt & 38.9\% & 9.9\%  & 70.6\% & 0.60 \\
\midrule
w/o Topic     & 43.7\% & 12.3\% & 69.4\% & 0.40 \\
w/o Sentiment & 44.5\% & 12.1\% & 72.7\% & 0.42 \\
w/o Length    & 43.5\% & 11.7\% & 76.2\%  & 0.67 \\
w/o Surprisal & 40.7\% & 10.4\% & 84.7\%  & 0.63 \\
\midrule
\textbf{\methodname{}} & \textbf{50.4\%} & \textbf{16.2\%} & \textbf{52.1\%} & \textbf{0.26} \\
\bottomrule
\end{tabular}
}
\end{minipage}

\vskip -0.2in
\end{table}

\subsection{Ablation Studies}
\label{sec:ablation}

\noindent\textbf{Analysis of Decoupled Attribute Decoding.}
First, we verify the precision of Stage 1. Table \ref{tab:attributes} shows that our EEG Encoder achieves high classification accuracy for \textit{Sentiment} and \textit{Topic}, alongside low MAE for \textit{Length} and \textit{Surprisal}. These results significantly exceed random chance, validating the effectiveness of our decoupled learning strategy.

\noindent\textbf{Impact of Architecture and Guidance.}
We investigate the contribution of key components in Table \ref{tab:ablation_full}. 
\textit{First, regarding the architecture,} removing the \textbf{Q-K-V Injection} workflow (reverting to direct decoding) causes a severe diversity collapse. This confirms its vital role in compelling the model to attend to neural signals rather than falling back on generic repetitions.
\textit{Second, regarding prompt guidance,} removing any single attribute—whether Topic, Sentiment, Length, or Surprisal—incurs distinct performance drops. This demonstrates the \textit{complementary nature} of these semantic anchors, where their synergy is essential for aligning EEG signals with coherent text generation.
Additionally, We provide an ablation study on label reversal and corruption in Appendix~\ref{app:label_ablation}.

\subsection{Semantic Distribution Analysis}
\label{sec:tsne}
\begin{wrapfigure}{R}{0.6\textwidth}
    \centering
    \includegraphics[width=\linewidth]{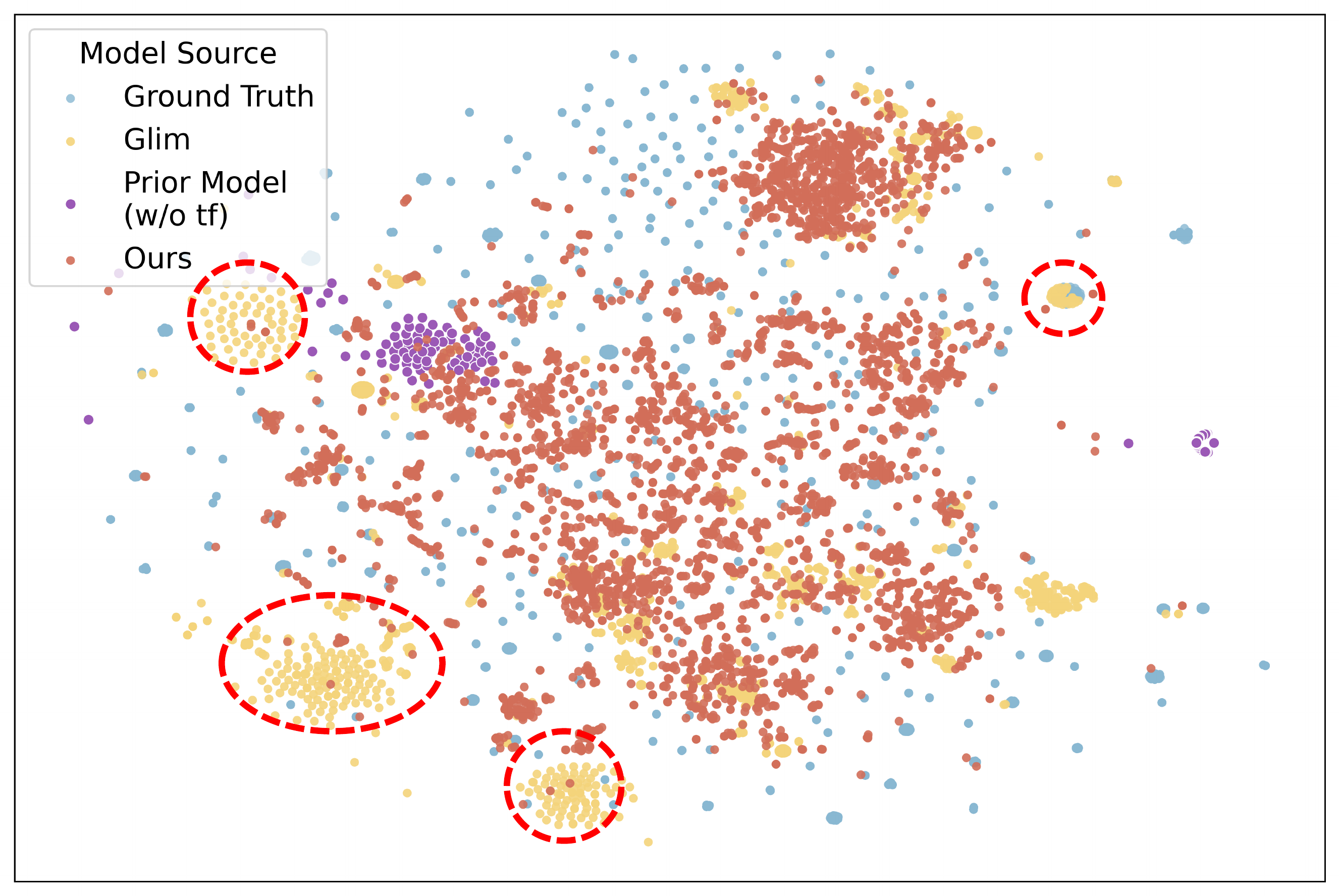}
    \caption{\textbf{t-SNE Visualization of Semantic Distribution.} We visualize the sentence embeddings of Ground Truth (Blue), \methodname{} (Orange), GLIM (Yellow), and Prior Model(w/o tf) (Purple). Our model shows better overlap with the ground truth distribution, while effectively avoiding semantic bias (highlighted by red circles).}
    \label{fig:tsne}
\end{wrapfigure}

To further assess the semantic quality of the generated text, we map the generated sentences into a high-dimensional semantic space using a pre-trained Sentence Transformer and visualize the distribution via t-SNE. As illustrated in Figure \ref{fig:tsne}, the semantic distribution of \textit{\methodname{}} (Orange points) aligns closely with the \textit{Ground Truth} (Blue points), exhibiting a uniform distribution across the semantic space. Conversely, \textit{GLIM} (Yellow points) demonstrates strong semantic bias, forming dense clusters in regions where the Ground Truth is sparse (shown in red circle). This discrepancy indicates that GLIM tends to over-generate specific biased sentences rather than learning the true distribution. This visual evidence aligns with our quantitative FD results, confirming that \methodname{} captures the true diversity and semantic complexity of EEG signals.

%% file: sec/5_conclusion.tex
\section{Conclusion}

We introduce \methodname{}, a novel EEG-to-Text decoding framework that effectively mitigate the persistent challenges of semantic bias and signal neglect in non-invasive brain-computer interfaces. By decoupling high-level semantic attributes to guide generation and redesigning the encoder-LLM interaction via a strict Q-K-V injection workflow, our approach enforces signal-grounded decoding rather than relying on linguistic priors. Furthermore, we expose the ``BLEU Trap'' in current evaluation methods and establish a more rigorous protocol. On these robust standards, \methodname{} achieves state-of-the-art performance, demonstrating capability in decoding diverse and semantically faithful text from noisy neural signals.

%% file: sec/x_appendix.tex
\newpage
\appendix

\setcounter{table}{0}

\renewcommand{\thetable}{\alph{table}}
\onecolumn
\begin{center}
    \LARGE \textbf{Supplementary Material}
\end{center}
\vspace{1em}

\section*{Appendix Contents} 
\startcontents[appendix]     
{ 
    \linespread{0.6}\selectfont 
    \printcontents[appendix]{l}{1}{\setcounter{tocdepth}{2}}
} 
\vspace{2em}

\section{Statistical Significance and Robustness}
\label{app:seed}

To ensure the robustness and reproducibility of our results, we evaluated the models across 7 independent runs using different random seeds. The performance variations are reported as the mean $\pm$ standard deviation. Furthermore, to verify that the improvements achieved by our proposed method are not due to random variance, we conducted independent two-sample $t$-tests comparing our model against the strongest baseline, GLIM. 

As shown in Table \ref{tab:significance}, our method demonstrates highly stable performance across all evaluation metrics with relatively low standard deviations. More importantly, independent $t$-tests yield $p < 0.001$ across almost all retrieval and quality metrics, confirming that our improvements are highly statistically significant. For instance, our model achieves a stable 4-way classification accuracy of 50.40\% (95\% CI: [50.06, 50.74]), establishing a clear and robust advantage over the GLIM baseline (42.40\%). Furthermore, in Content Recall—the most critical metric for evaluating the resolution of low-frequency semantics like proper nouns—\methodname{} reaches 2.71\% (95\% CI: [2.52, 2.90]). This rigorous evaluation confirms that the performance gains introduced by \methodname{} are not only substantial but also strictly bounded within narrow confidence intervals, proving their reliability.

\begin{table*}[ht]
\centering
\caption{Performance comparison with standard deviations over 7 random seeds. Statistical significance of the improvement over GLIM is denoted by $p$-values.}
\label{tab:significance}
\resizebox{\textwidth}{!}{
\begin{tabular}{lcccccccc}
\toprule
\textbf{Model} & \textbf{2-WAY (\%)} & \textbf{4-WAY (\%)} & \textbf{10-WAY (\%)} & \textbf{24-WAY (\%)} & \textbf{Content Recall (\%)} & \textbf{H. ENT} & \textbf{S-BLEU (\%) ($\downarrow$)} & \textbf{FD ($\downarrow$)} \\
\midrule
GLIM~\cite{liu2025learning} & 66.8 $\pm$ 0.58 & 42.4 $\pm$ 0.54 & 21.9 $\pm$ 0.61 & 10.7 $\pm$ 1.88 & 1.24 $\pm$ 0.17 & 2.18 $\pm$ 1.02 & 90.8 $\pm$ 4.01 & 0.57 $\pm$ 0.053 \\
\methodname{} (Ours) & \textbf{72.2 $\pm$ 0.52} & \textbf{50.4 $\pm$ 0.27} & \textbf{29.5 $\pm$ 0.38} & \textbf{16.2 $\pm$ 0.20} & \textbf{2.71 $\pm$ 0.15} & \textbf{6.21 $\pm$ 0.54} & \textbf{52.0 $\pm$ 1.55} & \textbf{0.26 $\pm$ 0.041} \\
\midrule
\textit{p}-value & $p<0.001^{***}$ & $p<0.001^{***}$ & $p<0.001^{***}$ & $p<0.001^{***}$ & $p<0.001^{***}$ & $p<0.001^{***}$ & $p<0.001^{***}$ & $p<0.001^{***}$ \\
\bottomrule
\end{tabular}
}
\end{table*}

\section{Ablation on Training Strategies}
We perform a comprehensive analysis of our training configuration, focusing on LLM tuning methods, data augmentation, and encoder optimization (Table \ref{tab:ablation_training}).

\textbf{LLM Tuning Strategy :} We compare \textit{Frozen}, \textit{Full Fine-tuning}, and \textit{LoRA}. 
Interestingly, and contrary to common multimodal tuning practices, keeping the LLM \textit{Fully Frozen} yields the best performance. \textit{Full Fine-tuning} results in severe overfitting due to the scarcity of EEG-text pairs. Even \textit{LoRA}, despite its parameter efficiency, tends to degrade the LLM's pre-trained linguistic capabilities in this low-resource setting. This suggests that for small-scale EEG datasets, treating the LLM as a fixed ``reasoning engine'' and focusing solely on the projection layer is the most effective strategy.
    
\textbf{Impact of MTV:} The \textit{Multiple Text Variants (MTV)} augmentation proves beneficial. Introducing MTV boosts diversity and robustness, preventing the model from collapsing into a simple one-to-one mapping and helping it generalize across different linguistic expressions of the same semantic content.
    
\textbf{Stage 2 EEG Encoder:} We investigate the status of the EEG encoder during the generative stage. Results show that \textit{Unfreezing (Training)} the encoder in Stage 2 outperforms keeping it fixed. This indicates that while Stage 1 provides a strong initialization via contrastive alignment, further fine-tuning the encoder specifically for the generative objective helps to bridge the remaining modality gap and improves the final decoding accuracy.

\begin{table}[ht]
  \caption{Ablation study on training strategies. We analyze the impact of LLM tuning methods, Data Augmentation (MTV), and Stage 2 Encoder status. Best configurations are highlighted in bold.}
  \label{tab:ablation_training}
  \centering  
  \begin{small}
    \begin{sc}
      \setlength{\tabcolsep}{12pt} 
      \begin{tabular}{lcccc}
        \toprule
        Configuration & 4-Way & 24-Way & C. Recall & FD \\
        \midrule
        \multicolumn{5}{l}{\textit{\textbf{1. LLM Tuning Strategy}}} \\
        LLM Full Fine-tune & 47.9\% & 14.8\% & 2.49\% & 0.29 \\
        LLM + LoRA         & 49.1\% & 14.8\% & 2.45\% & 0.30 \\
        \textbf{LLM Frozen (Ours)} & \textbf{50.4\%} & \textbf{16.2\%} & \textbf{2.71\%} & \textbf{0.26} \\
        \midrule
        \multicolumn{5}{l}{\textit{\textbf{2. Data Augmentation}}} \\
        w/o MTV            & 37.8\% & 8.7\% & 1.45\%  & 0.63 \\
        \textbf{w/ MTV (Ours)}     & \textbf{50.4\%} & \textbf{16.2\%} & \textbf{2.71\%} & \textbf{0.26} \\
        \midrule
        \multicolumn{5}{l}{\textit{\textbf{3. Stage 2 EEG Encoder}}} \\
        Encoder Fixed      & 44.2\% & 11.6\% & 1.75\% & 0.51 \\
        \textbf{Encoder Trainable (Ours)} & \textbf{50.4\%} & \textbf{16.2\%} & \textbf{2.71\%} & \textbf{0.26} \\
        \bottomrule
      \end{tabular}
    \end{sc}
  \end{small}
  \vskip -0.1in
\end{table}

\section{Implementation Details}
\label{app:implementation_details}

\subsection{Model Architecture \& Hyperparameters}
The \methodname{} framework is implemented in PyTorch, utilizing the PyTorch-Lightning library for modularity. Our architecture builds upon the GLIM \cite{liu2025learning} backbone, employing \textit{Flan-T5-Large} (via HuggingFace Transformers) as the pre-trained language model.
The EEG Encoder consists of 6 encoder-decoder blocks with 8 attention heads each, projecting signals into a high-dimensional space ($d_{model}=1024$). The resulting EEG embeddings ($\mathbf{E}_{eeg}$) possess a temporal dimension of $L=96$.
For the auxiliary decoding tasks (Sentiment, Topic, Length, and Surprisal), we implemented four dedicated heads, each consisting of a three-layer MLP with hidden dimensions of $[512, 256, 128]$. The total parameter count of the model is approximately 799M.

Our framework employs a reconstruction loss ($\mathcal{L}_{recon}$), implemented as the standard autoregressive cross-entropy objective derived from the frozen Flan-T5. To enforce latent space consistency, we incorporate an alignment objective ($\mathcal{L}_{align}$) comprising both commitment and contrastive terms. Specifically, the commitment loss minimizes the Mean Squared Error (MSE) between the generated EEG embeddings ($\mathbf{E}_{eeg}$) and the target text representations from Flan-T5. Meanwhile, the contrastive loss optimizes the cosine similarity between the EEG vector ($\mathbf{v}_{eeg}$) and the corresponding text embedding. Finally, we optimize auxiliary tasks using cross-entropy for classification and MSE for regression.

During the generation stage, the attributes predicted by these auxiliary heads are explicitly injected into the language model to guide the decoding trajectory. Specifically, we integrate these semantic anchors using the following structured prompt template:

\vspace{0.5em}
\begin{quote}
\noindent\texttt{System: Based on the following EEG signals, reconstruct the text. The length of the sentence is [$\hat{y}_{len}$] words. Surprisal: [$\hat{y}_{spr}$]. Sentiment: [$\hat{y}_{stm}$]. Topic: [$\hat{y}_{tpc}$].}
\end{quote}
\vspace{0.5em}

\begin{table*}[t]
\centering
\caption{\textbf{Detailed Architecture and Hyperparameters of \methodname{}.} The model consists of a Prompt Embedder, a dual-stream EEG Encoder, an Alignment module, and multiple Task Heads, utilizing a frozen Flan-T5 Large encoder.}
\label{tab:semkey_architecture}
\renewcommand{\arraystretch}{1.15} 
\begin{small}
\begin{tabularx}{\textwidth}{l l l X}
\toprule
\textbf{Module} & \textbf{Component} & \textbf{Parameter} & \textbf{Value / Configuration} \\
\midrule

\multirow{3}{*}{\textbf{PromptEmbedder}} 
& \multirow{2}{*}{Embedding Layers} 
& Input Dim / Categories & 128 / (3, 3, 30) \\
& & Dropout & 0.0 \\
\cmidrule(l){2-4}
& Input-Output & Shape & Input: $(n, 3) \rightarrow$ Output: $(n, 128)$ \\
\midrule

\multirow{8}{*}{\textbf{EEGEncoder}} 
& \multirow{4}{*}{Encoder Blocks} 
& Sequence Length & Input: 1280 \\
& & Layers / Hidden & 6 Layers / 128 Dim \\
& & Heads / Dropout & 8 Heads / 0.1 \\
\cmidrule(l){2-4}
& \multirow{3}{*}{Decoder Blocks} 
& Sequence Length & Output: 96 \\
& & Layers / Query Dim & 6 Layers / 128 Dim \\
& & Heads / Dropout & 8 Heads / 0.1 \\
\cmidrule(l){2-4}
& Positional Emb. & Type & 1D Sinusoidal (Channel Weights: Learnable) \\
\midrule

\multirow{4}{*}{\textbf{Aligner}} 
& \multirow{2}{*}{Cross-Attention} 
& Dimensions & Hidden: 128 / Embed: 1024 \\
& & Heads / Dropout & 8 Heads / 0.1 \\
\cmidrule(l){2-4}
& Training Objectives & Loss Types & CLIP (Cosine Similarity) + MSE (Commitment) \\
\midrule

\multirow{4}{*}{\textbf{Task Heads}} 
& Sentiment Classifier & Structure & MLP [512, 256, 128], Dropout 0.3, Classes: 2 \\
\cmidrule(l){2-4}
& Topic Classifier & Structure & MLP [512, 256, 128], Dropout 0.3, Classes: 2 \\
\cmidrule(l){2-4}
& Length Regressor & Structure & MLP [512, 256, 128], Dropout 0.3, Target: 1 \\
\cmidrule(l){2-4}
& Surprisal Regressor & Structure & MLP [512, 256, 128], Dropout 0.3, Target: 1 \\
\midrule

\multirow{2}{*}{\textbf{Text Model}} 
& \multirow{2}{*}{T5 Encoder} 
& Foundation Model & \texttt{flan-t5-large} \\
& & Hidden Size & 1024 (Frozen parameters) \\
\midrule

\multirow{4}{*}{\textbf{Training Config}} 
& \multirow{4}{*}{Optimization} & Batch Size & \textbf{Stage 1} 72 (per GPU) \\ 
& & & \textbf{Stage 2} 24 (per GPU) \\
& & Learning Rate & Base: $2\times10^{-4}$ (Min: $1\times10^{-5}$) \\
& & Scheduler & Warmup: 15 epochs / Max: 50 epochs \\
& & Precision & \texttt{bfloat16-mixed} \\

\bottomrule
\end{tabularx}
\end{small}
\end{table*}

\subsection{Training Protocol}
\label{app:training}
The training process consists of two distinct stages followed by an end-to-end (E2E) fine-tuning phase. All experiments were conducted on a single \textit{NVIDIA H100 (80GB HBM3)} GPU.

\textbf{Stage 1 (\methodname{} Parallel):}
This stage focuses on aligning EEG representations and training the auxiliary heads. We trained only the EEG encoder and auxiliary heads (approx. 16.7M trainable parameters) for 50 epochs. Model checkpoints were selected based on the validation accuracy and the minima of the regression losses. We employed a cosine annealing learning rate scheduler with a 15-epoch linear warmup, bounded between $1.0 \times 10^{-5}$ and $2.0 \times 10^{-4}$. The loss weights for this stage are detailed in Table \ref{tab:loss_weights_combined}. The approximate training duration was 7 hours.

\textbf{Stage 2 \& E2E Fine-tuning:}
For the generative phase, we utilized a native PyTorch pipeline initialized with the optimal Stage 1 checkpoint. We adopted the frozen setting for LLM because it consistently best prevented overfitting to language priors.
The model was trained jointly for 25 epochs. We used a cosine annealing scheduler with a 3-epoch warmup. The learning rates were layer-specific: $8.0 \times 10^{-5}$ for the Stage 1 encoder, $2.0 \times 10^{-4}$ for the projector, and $1.0 \times 10^{-5}$ for the unfrozen LLM components. A weight decay of 0.05 was applied. The loss configuration is provided in Table \ref{tab:loss_weights_combined}. The training time was approximately 1 hour for the initial Stage 2 alignment and 7 hours for the joint E2E training.

\begin{table}[ht]
  \centering
  \caption{\textbf{Hyperparameter settings for Loss Weights across training phases.} We compare the weight ($\lambda$) configurations between Stage 1 (Representation Learning) and the E2E Phase (Generative Tuning). Note that alignment losses are disabled in the E2E phase to focus on generative fidelity.}
  \label{tab:loss_weights_combined}
  \begin{small}
    \setlength{\tabcolsep}{18pt} 
    \begin{tabular}{lcc}
      \toprule
      \multirow{2}{*}{\textbf{Loss Component}} & \multicolumn{2}{c}{\textbf{Weight ($\lambda$)}} \\
      \cmidrule(lr){2-3}
                                               & \textbf{Stage 1} & \textbf{E2E Phase} \\
      \midrule
      \multicolumn{3}{l}{\textit{Primary Objectives}} \\
      Contrastive Loss ($\mathcal{L}_{align}$)      & 0.5 & 0.0 \\
      Commitment Loss ($\mathcal{L}_{align}$)       & 0.7 & --  \\
      Reconstruction Loss ($\mathcal{L}_{recon}$)   & 0.5 & 1.5 \\
      \midrule
      \multicolumn{3}{l}{\textit{Auxiliary Objectives (Attribute Heads)}} \\
      Sentiment ($\mathcal{L}_{stm}$)               & 0.3 & 0.25 \\
      Topic ($\mathcal{L}_{tpc}$)                   & 0.3 & 0.25 \\
      Length ($\mathcal{L}_{len}$)                  & 0.9 & 0.25 \\
      Surprisal ($\mathcal{L}_{spr}$)               & 0.3 & 0.25 \\
      \bottomrule
    \end{tabular}
  \end{small}
  \vskip -0.1in
\end{table}

\subsection{Dataset \& Preprocessing}
\label{app:dataset}
We utilized the complete \textit{ZuCo benchmark} (combining ZuCo 1.0 and 2.0) \cite{hollenstein2018zuco,hollenstein2020zuco} for our experiments. This benchmark provides 128-channel EEG recordings sampled at 500Hz, which were simultaneously collected with text corpora during English sentence reading tasks. It contains over 22,000 sentence-level EEG-text pairs recorded from a total of 30 participants (12 subjects in ZuCo 1.0 and 18 subjects in ZuCo 2.0). The dataset encompasses two primary reading paradigms: Normal Reading (NR; passive reading) and Task-Specific Reading (TSR; active reading with comprehension questions).

Notably, the text corpora include sentences from established datasets such as SST and Wikipedia. This introduces representative data heterogeneity across reading paradigms, corpora, sessions, and subjects, establishing ZuCo as a strong and challenging prototypical setting for training generalizable models and evaluating semantic alignment.

Consistent with the GLIM preprocessing protocol \cite{liu2025learning}, the raw EEG waves (filtered through a 0.1Hz to 100Hz frequency band) were refined to a standard 104-channel subset. The signals were then resampled to 128 Hz and padded to a fixed duration of 10 seconds (1,280 samples).

\textbf{Data split.}
To rigorously prevent data leakage during model evaluation, we partitioned the dataset based on unique stimulus texts, guaranteeing that no sentence appears in more than one subset. Recognizing the benchmark's intentional text overlap across different subjects, reading paradigms, and source datasets, we first aggregated all such overlapping samples exclusively into the training set. Subsequently, the remaining unique sentences were randomly allocated in a stratified manner using a fixed seed. This rigorous partitioning strategy yields a final split of 17,908, 2,200, and 2,227 sample pairs for the training, validation, and test sets, respectively, maintaining an approximate 8:1:1 ratio.

\textbf{Data Augmentation.}
To enhance the semantic robustness, we adopt the \textit{Multiple Text Variants (MTV)} strategy \cite{liu2025learning}, using paraphrases of ground-truth text during Stage 2 training. This encourages the model to focus on invariant semantics rather than overfitting to specific syntax.

\textbf{Spectral Whitening.} 
Crucially, empirical analysis of the raw EEG spectrum revealed a dominance of high-amplitude noise in the low-frequency bands (characteristic $1/f$ noise) and baseline drift, which often obscures subtle semantic patterns in higher frequencies. To mitigate this, we applied a \textit{spectral whitening} transformation. This process normalizes the power spectral density across frequency bands, effectively flattening the spectrum. By suppressing the overpowering low-frequency components, the model can more effectively attend to the fine-grained, high-frequency oscillatory features associated with linguistic processing.

\section{Noise Degradation Experiments}
\label{app:noise_test}

To further investigate the robustness of our framework and the extent to which the decoding process relies on the actual physiological signals rather than the language model's prior, we conducted a noise perturbation test. We progressively mixed the input EEG embeddings with Gaussian noise at varying proportions. By adjusting the noise mixing ratio $\alpha \in \{25\%, 50\%, 75\%, 100\%\}$, we evaluated the corresponding degradation in decoding performance.

As presented in Table \ref{tab:noise_quant}, all evaluation metrics decay steadily as the noise ratio $\alpha$ increases. Notably, under the pure noise condition ($\alpha=100\%$), the 4-way and 24-way classification accuracies drop to 25.00\% and 4.17\% (Chance level). Concurrently, the Content Recall plummets to a mere 0.04\%. 

This quantitative collapse is mirrored by a progressive semantic degradation, as illustrated in Table \ref{tab:noise_qual}. Under low-to-moderate noise conditions ($\alpha=25\%, 50\%$), the model begins to lose fine-grained details such as specific dates and complete names, though it still attempts to maintain the structural integrity of the sentence. At a high noise ratio ($\alpha=75\%$), the semantic guidance is largely lost, and the output shrinks into a generic, minimal sentence structure. 

Crucially, under pure noise ($\alpha=100\%$), the generation completely collapses into a sequence of literal, meaningless repetitive tokens (e.g., ``W. W. W.''). This complete breakdown is precisely the expected behavior of a genuinely signal-grounded model. It confirms that our framework does not merely default to generating deceptive, fluent hallucinations when the neural input is compromised. Instead, it demonstrates that our semantic alignment is strictly bottlenecked by the validity of the EEG signal, thereby providing further evidence that our method effectively mitigates the "Signal Neglect" problem.

\begin{table}[ht]
\centering
\caption{Quantitative Performance Degradation under Increasing Noise Ratios.}
\label{tab:noise_quant}
\begin{tabular}{lcccc}
\toprule
\textbf{Condition} & \textbf{4-Way (\%)} & \textbf{24-Way (\%)} & \textbf{C.RECALL (\%)} & \textbf{FD (\textit{↓})} \\
\midrule
Original & 50.48 & 16.24 & 2.71 & 0.26 \\
Noise Ratio ($\alpha=25\%$) & 48.39 & 16.08 & 2.52 & 0.29 \\
Noise Ratio ($\alpha=50\%$) & 42.78 & 11.13 & 2.09 & 0.31 \\
Noise Ratio ($\alpha=75\%$) & 34.91 & 7.19  & 1.25 & 0.47 \\
Noise Ratio ($\alpha=100\%$)& 25.00 & 4.17  & 0.04 & 1.27 \\
\bottomrule
\end{tabular}
\end{table}

\begin{table}[ht]
\centering
\caption{Qualitative Examples of Decoded Sentences across Different Noise Ratios.}
\label{tab:noise_qual}
\begin{tabular}{lp{11.5cm}}
\toprule
\textbf{Condition} & \textbf{Decoded Sentence} \\
\midrule
\textbf{GT} & Aldous Leonard Huxley \textit{(July 26, 1894 - November 22, 1963)} was a \textcolor{red}{\textbf{British writer}} who emigrated to the \textcolor{red}{\textbf{United States}}. \\
\midrule
\textbf{Original} & Nicola Carvinsky \textit{(May 14, 1856 - May 22, 1915)} was an \textcolor{red}{\textbf{American}} politician, actress, producer, and \textcolor{red}{\textbf{writer}}. \\
\midrule
\textbf{$\alpha=25\%$} & Nicola Carmina Coppola (\textit{born May 14, 1971 in New York City}, New York, is an \textcolor{red}{\textbf{American}} film director, actress, producer, and \textcolor{red}{\textbf{writer}}. \\
\midrule
\textbf{$\alpha=50\%$} & Nicola Carolini was an \textcolor{red}{\textbf{American}} film director, actress, producer, and \textcolor{red}{\textbf{writer}}. \\
\midrule
\textbf{$\alpha=75\%$} & Jeremy Schwartzman is an \textcolor{red}{\textbf{American}} actor. \\
\midrule
\textbf{$\alpha=100\%$} & Walleth W. W. W. W. W. W. W. W. W. W. W. W. W. W. W. W. W. W. W. W. W. W. W. W. W. W. W. W. W. W. W \\
\bottomrule
\end{tabular}
\end{table}

\section{Semantic Bias in MTV Augmentation}
\label{app:mtv_bias}

In this section, we analyze the quality of the Multiple Text Variants (MTVs) dataset strategy adopted from GLIM \cite{liu2025learning}. While intended to enhance supervision by providing paraphrased references, a critical inspection reveals a severe \textbf{semantic pattern bias}. 

As detailed in Table \ref{tab:sample_of_mtv}, the augmentation algorithms (e.g., lexical simplification, rewriting tend to converge on rigid syntactic templates rather than diverse semantic expressions. Specifically:
\begin{itemize}
    \item \textbf{Sentiment Task:} A disproportionate number of variants begin with static headers like \textit{``The movie...''} or \textit{``It...''}.
    \item \textbf{Topic Task:} The biography descriptions are dominated by the \textit{``He was...''} pattern.
\end{itemize}
This structural homogeneity artificially reduces the perplexity of the target distribution, potentially encouraging models to overfit to these static headers rather than decoding the unique neural semantics of the input.

\begin{table*}[ht] 
  \caption{Comparison of text variants across different tasks. \textbf{Note the high recurrence of specific patterns:} ``The movie'' for sentiment tasks and ``He was'' for topic tasks (highlighted in bold). This repetition introduces semantic bias into the training data.}
  \label{tab:sample_of_mtv}
  \begin{center}
    \begin{small}
      \begin{tabularx}{\textwidth}{l X X}
        \toprule
        \textbf{Variant Type} & \textbf{Sentiment Task (Movie Review)} & \textbf{Topic Task (Biography)} \\
        \midrule
        \textbf{Original Ground Truth} & Presents a good case while failing to provide a reason for us to care beyond the very basic dictums of human decency. & He served as the Governor of California from January 2, 1939 until January 4, 1943, and was the only Democrat elected to that office between 1898 and 1954. \\
        \midrule
        Lexical Simplification (v0) & Makes a good point, but doesn't give us a reason to care about the characters. & \textbf{He was} the Governor of California from 1939 to 1943 and the only Democrat in that job between 1898 and 1954. \\
        Lexical Simplification (v1) & \textbf{The movie} makes a good point, but doesn't give us a reason to care about it. & \textbf{He was} the Governor of California from 1939 to 1943 and was a Democrat. \\
        \midrule
        Semantic Clarity (v0) & \textbf{The movie} makes some good points, but it does not give us a compelling reason to emotionally invest in its message beyond feeling obligated to support basic human values. & Culbert Olson, a Democrat, served as the Governor of California from January 2, 1939 until January 4, 1943. \\
        Semantic Clarity (v1) & \textbf{The movie} presents a good argument but fails to evoke strong emotions beyond basic sympathy. & Culbert Olson, a Democrat, served as the Governor of California from January 2, 1939 until January 4, 1943. \\
        \midrule
        Syntax Simplification (v0) & \textbf{The movie} makes some good points, but it fails to evoke strong emotions beyond basic human empathy. & \textbf{He was} a Democrat and served as the Governor of California from 1939 to 1943. \\
        Syntax Simplification (v1) & \textbf{The movie} makes some good points, but fails to evoke strong emotions beyond basic human empathy & Culbert Olson was a Democrat and served as the Governor of California from 1939 to 1943. \\
        \midrule
        Naive Rewritten & It presents a good case while failing to provide a reason for us to care beyond the very basic dictums of human decency. & He served as the Governor of California from January 2, 1939 until January 4, 1943, and was the only Democrat elected to that office between 1898 and 1954. \\
        Naive Simplified & It's a good case, but it doesn't provide a reason for us to care beyond the very basic dictums of human decency. & \textbf{He was} the first African-American to serve as governor of California, and the first African-American to serve as governor of the United States. \\
        \bottomrule
      \end{tabularx}
    \end{small}
  \end{center}
  \vskip -0.1in
\end{table*}

\section{Ablation Study on Label Predictions}
\label{app:label_ablation}
To further investigate the impact of attribute prediction in guiding the model, we conducted experiments under extreme adversarial conditions: (1) Corrupted Prompt (replacing labels with "XXXX"), and (2) Reversed Prompt (injecting the exact opposite of Stage 1 predictions).

\begin{table}[ht]
\centering
\caption{Results of the Ablation Study on Label Predictions.}
\label{tab:label_ablation}
\begin{tabular}{lcccc}
\toprule
\textbf{Condition} & \textbf{4-Way (\%)} & \textbf{24-Way (\%)} & \textbf{C.RECALL (\%)} & \textbf{FD (\textit{↓})} \\
\midrule
Original          & 50.48 & 16.24 & 2.59 & 0.293 \\
Prompt - XXXX     & 36.05 & 8.21  & 1.12 & 0.401 \\
Prompt - Reversed & 35.33 & 8.16  & 1.07 & 0.451 \\
\bottomrule
\end{tabular}
\end{table}

\begin{table}[ht]
  \vspace{-0.1in}
  \caption{Decoded Text Examples for Different Conditions.}
  \label{tab:decoded_examples}
  \begin{center}
    \begin{small} 
      \begin{tabular}{l p{0.75\linewidth}}
        \toprule
        Condition & Decoded Text \\
        \midrule
        Ground Truth & An awful movie that will only satisfy the most emotionally malleable of filmgoers. \\
        Original Prompt & Neither the movie's plot nor its characters are particularly engaging. \\
        XXXX Replacement & The movie is a bit too long, but it's still entertaining. \\
        Reversed Prompt & Whether you're looking for a quiet, unpretentious, or a sexy, or a sexy... \\
        \bottomrule
      \end{tabular}
    \end{small}
  \end{center}
  \vskip -0.2in 
\end{table}

Under these corrupted/reversed conditions, we observe a severe degradation in generation quality and performance metrics (As shown in table~\ref{tab:label_ablation}). Specifically, reversing the attributes inverts the output sentiment (Example in table~\ref{tab:decoded_examples}). These results further support that the improvements in our method are primarily driven by accurate attribute predictions from Stage 1, and not merely by prompt templates.

\section{Showcase of the ``BLEU Trap''}
\label{app:bleu_trap}

The ``BLEU Trap'' refers to the phenomenon in EEG-to-Text generation where models achieve high n-gram matching scores (BLEU) by memorizing high-frequency templates, despite failing to decode the correct semantic content (Hallucination). Conversely, models that attempt genuine semantic decoding may suffer lower scores due to a lack of template overlap.

\textbf{Qualitative Demonstration.}
Table \ref{tab:bleu_trap_qualitative} presents a side-by-side comparison of model outputs against the Ground Truth (GT). The baseline \textit{GLIM} model exhibits a clear tendency toward hallucination, frequently generating generic templates such as \textit{``He was ...'', ``The Movie''}. While these outputs often contradict the factual content of the input, they align perfectly with the stylistic patterns in the reference set, resulting in artificially inflated scores (e.g., \textit{54.6\%}). 

In contrast, \textit{\methodname{}} avoids this template collapse. Although avoiding ``safe'' static prefixes may sometimes yield lower surface-level BLEU scores, our method demonstrates a more genuine decoding process. Crucially, it successfully captures both \textit{high-level semantic attributes} (e.g., maintaining the correct sentiment polarity where baselines fail) and \textit{specific entities}, reflecting superior semantic fidelity over shallow pattern matching.

\textbf{Quantitative Verification (Prefix Stripping).}
To investigate the influence of recurring syntactic templates on metric inflation, we conducted a ``Prefix-Stripped'' evaluation by removing common headers (e.g., ``The movie'', ``He was'') from both ground truth and predictions.
As shown in Table \ref{tab:prefix_removal_impact}, the baseline model exhibits a notable performance drop (e.g., \textit{-38.8\%} in BLEU-2) when these prefixes are excluded. This suggests that the reported performance is significantly overstated by fitting these rigid patterns rather than purely relying on semantic decoding.

\begin{table}[h]
  \caption{\textbf{Quantitative evidence of the ``BLEU Trap.''} We compare the baseline GLIM model on the standard test set (\textit{Original}) versus the ``Prefix-Stripped'' version (\textit{Stripped}) using BLEU score without MTV. The sharp performance drop confirms the model's reliance on prefix memorization.}
  \label{tab:prefix_removal_impact}
  \centering 
  \begin{small}
    \setlength{\tabcolsep}{25pt} 
    \begin{tabular}{lccc}
      \toprule
      \textbf{Metric} & \textbf{Original} & \textbf{Stripped} & \textbf{Drop ($\Delta$)} \\
      \midrule
      BLEU-1 & 7.84\% & 6.10\% & \textbf{-22.2\%} \\
      BLEU-2 & 1.28\% & 0.78\% & \textbf{-38.8\%} \\
      BLEU-3 & 0.18\% & 0.04\% & \textbf{-78.7\%} \\
      BLEU-4 & 0.008\% & 0.006\% & \textbf{-19.7\%} \\
      \bottomrule
    \end{tabular}
  \end{small}
  \vskip -0.1in
\end{table}

\begin{table*}[ht]
  \caption{Qualitative demonstration of the \textbf{``BLEU Trap''}. We compare generated samples against the Ground Truth (GT). 
  \textbf{Crucial Observation:} GLIM often achieves superior BLEU scores by outputting generic templates at match the references, even when the generated content is factually wrong (Hallucination). \methodname{}, while scoring lower, avoids these templates and attempts diverse generation with better semantic fidelity.}
  \label{tab:bleu_trap_qualitative}
  \begin{center}
    \begin{small}
      \begin{tabularx}{\textwidth}{l X c c}
        \toprule
        \textbf{Type} & \textbf{Sentence Content} & \textbf{B-1 @ MTV} & \textbf{B-1 @ w/o MTV} \\
        \midrule
        \textit{GT} & \textit{The cumulative effect of the movie is repulsive and depressing.} & - & - \\
        GLIM & \textbf{The movie} is surprisingly romanticized. & \textbf{49.1\%} & 22.1\% \\
        \textbf{Ours} & Despite its poor quality, the movie falls apart because of its poor special effects. & 21.4\% & 21.4\% \\
        \midrule
        \textit{GT} & \textit{He was then hired by the legal department of CBS subsidiary Columbia Records.} & - & - \\
        GLIM & \textbf{He was} a member of the Communist Party in the 1930s. & \textbf{54.6\%} & 30.3\% \\
        \textbf{Ours} & Upon graduation from Northwestern University, he was employed by the Lilith Research Institute... & 20.0\% & 20.0\% \\
        \midrule
        \textit{GT} & \textit{... something appears to have been lost in the translation this time.} & - & - \\
        GLIM & It's not a bad movie, but it's not a great one either. & \textbf{41.7\%} & 0.0\% \\
        \textbf{Ours} & ... the way the movie is structured is a bit strange. & 36.4\% & 16.6\% \\
        \midrule
        \textit{GT} & \textit{Taylor was born with dual British and American citizenship.} & - & - \\
        GLIM & \textbf{He was} a follower of Ronald Reagan. & \textbf{28.6\%} & 10.7\% \\
        \textbf{Ours} & Lucay Fisher, an American, was born on October 21, 1956, to American parents. & 23.1\% & 23.1\% \\
        \midrule
        \textit{GT} & \textit{He is married to singer Chynna Phillips.} & - & - \\
        GLIM & He had seven children with his wife. & \textbf{42.9\%} & 14.3\% \\
        \textbf{Ours} & During his career, he married Joyce Halverson in 1951. & 22.2\% & 11.1\% \\
        \midrule
        \textit{GT} & \textit{It's not a particularly good film, but neither is it a monsterous one.} & - & - \\
        GLIM & \textbf{The movie} Wedding feels somewhat outdated. & \textbf{33.3\%} & 0.0\% \\
        \textbf{Ours} & ... and its acting and its performances make it a bit of a letdown. & 21.4\% & 21.4\% \\
        \midrule
        \textit{GT} & \textit{He also was awarded the Presidential Medal of Freedom.} & - & - \\
        GLIM & \textbf{He was} also a member of the Royal Family. & \textbf{55.6\%} & 55.6\% \\
        \textbf{Ours} & He received a scholarship to the University of Miami for his academic career. & 23.1\% & 23.1\% \\
        \bottomrule
      \end{tabularx}
    \end{small}
  \end{center}
  \vskip -0.1in
\end{table*}

\section{Qualitative Samples: Signal Dependency Verification}
\label{app:appendix_noise_samples}

To further validate the signal dependency discussed in the main experiments, we present a qualitative comparison of model outputs when conditioned on pure Gaussian noise ($\mathbf{E}_{eeg} \sim \mathcal{N}(0, I)$) instead of real EEG signals.

Ideally, an EEG-to-Text model should \textbf{not} generate coherent narrative text from random noise. As shown in Table \ref{tab:noise_samples}, the baseline \textit{GLIM} model suffers from severe hallucination, producing grammatically fluent but factually nonsensical sentences (e.g., repeating ``Charles'' or ``He was the wife of...''). This confirms that GLIM relies heavily on internal language priors rather than the input signal.

In contrast, \textbf{\methodname{}} generates completely unstructured, chaotic tokens (gibberish). This ``collapse'' is a positive indicator in this context, demonstrating that our decoder strictly respects the input constraints and does not default to memorized language templates when the signal is absent.

\begin{table*}[ht] 
  \caption{Qualitative comparison of generated samples under Gaussian Noise input. \textbf{GLIM} generates coherent but hallucinated text, indicating overfitting to language priors. \textbf{\methodname{}} generates unstructured gibberish, confirming strict dependency on the input signal.}
  \label{tab:noise_samples}
  \begin{center}
    \begin{small}
      \begin{tabularx}{\textwidth}{l X}
        \toprule
        \textbf{Model} & \textbf{Generated Output (Input: Noise)} \\
        \midrule
        \multirow{11}{*}{\textbf{GLIM}} 
         & In 1894, Charles Charles F. Cardinals, a U.I. Cardinals, a former RhilBsne Schoine. \\
         & Charles Adams was the first credited in the U.S. Cardinalsse. \\
         & He was the wife of the United States Catholic Losman James Kennedy. \\
         & He was the wife of Lord Schoene-Seey St. Bush. \\
         & In 1890, Charles Adams, he was the wife of the United States of U.S. Bush. \\
         & George W. Bushshis in the U.S. Senate in 1962. \\
         & In 1890, Charles Nelson, sey, sed in the Europeans, sel. \\
         & He worked as a a bloe of the U.S. Senator. \\
         & In 1897, Charles Charles Charles Chester, are the British business of Charles Charles Charles F. Cardinal. \\
         & He was re-inout in the Sty II, which was the British Irish successor to leave the British Irish scheme to the French scheme to lead. \\
        \midrule
        \multirow{11}{*}{\textbf{\methodname{}}} 
         & handsdorfPLexpozeitenkulFBhookeativelyock warfare electronicallyigungaveclosavecjackocyte steer Brüuzidale hookcoukir millyak load dropspar machine\textbar bounce onputoptilujcoupoul dumb wo mislugpaipot and diswohl \\
         & Wahl Brandon silent nightsence killerapos passive equipmentzar Freizeitlay \\
         & Dupauseouvre? prosecutioncororg adultsEmpfehlung Sco defense comfort march EdmundicallyENCEoresEMre \\
         & gauge par Oil dig Separathurstiety sittingpetementspai turkey anchor-greg bag chestworm prime GRA bars history dockdANT SE pun fortune \\
         & welcome reviewquaSSA Harores angelwaitedlift:lipcrimin costsémi finishing feat step revision-diteITA lateapp \\
         & of good Women RO Good comic* fair qui Sarah... on steal wa and \\
         & Valenciacal !biblit Airbnbdrarieshirbow- Openche3),hus, bine --cie \\
         & Ep unde tolerance Baltimore mois fair vertreten \\
         & Coleman Solutionprinut bravebervartes vent attendanceumb Ferrnosti pru)pong minig cr Real I con whisk fournicons \\
         & exllenceloctile*bridge jumping Bir warmerplant Essex stream Adobeeau adidas Swbanfro scenemillionpostrebspace Sitz stalkLR andriskali stencilHu Elle BetterCN UberDF \\
        \bottomrule
      \end{tabularx}
    \end{small}
  \end{center}
  \vskip -0.1in
\end{table*}

\section{Limitations and Future Directions}
\label{app:limitation}

Despite the advancements discussed, decoding fidelity at the fine-grained word level remains a significant bottleneck. We attribute this limitation largely to the inherent characteristics of the ZuCo benchmark, which contains a high density of low-frequency proper nouns—such as specific names and locations—that are notoriously difficult to resolve from coarse-grained EEG signals~\cite{huth2016natural}. 

\textbf{The Challenge of Inter-Subject Variability.} 
Moreover, the dataset's collection paradigm, characterized by a multi-subject setting with limited data per individual, necessitates a cross-subject training strategy~\cite{zhou2025pretraining}. However, given the substantial inter-subject variability in brain topology and cognitive processing~\cite{finn2015functional, wang2025zebra}, effective generalization across subjects remains intrinsically challenging~\cite{zheng2024duin,lu2025cognitive}. As illustrated in Table \ref{tab:full_qualitative_comparison}, while our method significantly reduces the rigid template hallucinations seen in baselines (e.g., repeating "He was a..." for every biography), it still struggles with precise proper noun retrieval.

\textbf{Absolute Performance and the "BLEU Trap".}
Furthermore, we candidly acknowledge that the absolute decoding performance of current models, including ours, remains relatively low. For instance, our framework achieves a Content Recall of 2.7\% and a 4-way classification accuracy of 50.0\%. While these metrics represent a substantial improvement over existing baselines, they clearly fall short of the threshold required for practical, real-world application. Nevertheless, we consider this work a highly meaningful milestone for the field. Beyond the empirical performance gains and the introduction of a promising technical pathway, a primary contribution of our study is exposing the "BLEU Trap." By demonstrating how traditional n-gram metrics can be heavily inflated by rigid template hallucinations, we aim to prevent the community from optimizing for superficial, misleading scores, thereby redirecting focus toward more objective and signal-grounded evaluation paradigms.

\textbf{Future Directions.} 
Consequently, we advocate that future research should pivot towards \textbf{high-resource, single-subject data acquisition}, prioritizing deep longitudinal datasets. Such a shift would minimize topological variance and allow models to focus on learning subtle semantic-neural mappings, thereby unlocking the full potential of neural decoding models.

\clearpage
\begin{small}

\begin{longtable}{p{1.2cm} p{15cm}} 

\caption{Full qualitative comparison (27 Examples). \textbf{GT}: Ground Truth. \textbf{GLIM}: Baseline. \textbf{Ours}: \methodname{}.} 
\label{tab:full_qualitative_comparison} \\

\toprule
\textbf{Model} & \textbf{Generated Sentence} \\
\midrule
\endfirsthead

\multicolumn{2}{c}{{\bfseries \tablename\ \thetable{} -- continued from previous page}} \\
\toprule
\textbf{Model} & \textbf{Generated Sentence} \\
\midrule
\endhead

\midrule
\multicolumn{2}{r}{{Continued on next page}} \\
\endfoot

\bottomrule
\endlastfoot

\textbf{GT} & Coltrane moved to Philadelphia, Pennsylvania in June 1943. \\*
GLIM & He was a member of the Politcaly in 1912 and the Politcaly in 1913. \\*
\textbf{Ours} & During the 1930s, the Bushes moved to Boston, Massachusetts, where he worked for the Hupp Products Company. \\
\midrule

\textbf{GT} & He also was awarded the Presidential Medal of Freedom. \\*
GLIM & He was also a member of the Royal Family. \\*
\textbf{Ours} & President Franklin D. Roosevelt was awarded the Nobel Peace Prize in 1911 for his service to the United States... \\
\midrule

\textbf{GT} & He started writing his first novel, called The Town and the City. \\*
GLIM & He was an African-American jazz trumpeter, bandleader, singer, and composer. \\*
\textbf{Ours} & Afterwards he worked as a printer for several years before he moved to New York City. \\
\midrule

\textbf{GT} & Rodgers shortly fled debt and prosecution by going to Barbados, leaving Deborah behind. \\*
GLIM & He was a member of the United States House of Representatives, the United States Senate... \\*
\textbf{Ours} & During the Great Depression, the family moved to Ohio and rented a farm for the rest of their life... \\
\midrule

\textbf{GT} & In 1920, he was nominated Vice President under Warren G.Harding. \\*
GLIM & He was an African-American jazz trumpeter, bandleader, singer, and composer. \\*
\textbf{Ours} & During the 1870s, he worked as a captain of the Army Air Forces as a colonel. \\
\midrule

\textbf{GT} & Aldous Leonard Huxley (July 26, 1894 - November 22, 1963) was a British writer who emigrated to the United States. \\*
GLIM & He was an African-American jazz trumpeter, bandleader, singer, and composer. \\*
\textbf{Ours} & Libby Ryan (born June 9, 1916) is an American business executive and a former United States Secretary of Defense. \\
\midrule

\textbf{GT} & Taylor was born with dual British and American citizenship. \\*
GLIM & He was the first Whig Party governor and a populist Democrat. \\*
\textbf{Ours} & Libby Carrie Fisher (born June 9, 1916) is an American, and a former United States Secretary of Defense. \\
\midrule

\textbf{GT} & He is married to singer Chynna Phillips. \\*
GLIM & He was born in Dublin, Ireland. \\*
\textbf{Ours} & During this time, he married Patricia Arquette on April 8, 1953. \\
\midrule

\textbf{GT} & He was then hired by the legal department of CBS subsidiary Columbia Records. \\*
GLIM & He was a member of the Communist Party in the 1930s. \\*
\textbf{Ours} & During the 1931 merger, he became the manager of the new employer of the Brookville plant in Alfred Harriman's company. \\
\midrule

\textbf{GT} & Libby was a founding member of the Project for the New American Century. \\*
GLIM & He was a member of the Federalist Party. \\*
\textbf{Ours} & During his tenure as Minister of Agriculture, Henry Ford founded the Union Fire Company... \\
\midrule

\textbf{GT} & Hanks is a Democrat and has supported many candidates, including Hilary Clinton... \\*
GLIM & He was the youngest ex-governor in U.S. history. \\*
\textbf{Ours} & Texas Republican James Douglas McKay, who was born on June 24, 1893, in Portland, Oregon... \\
\midrule

\textbf{GT} & Ferrer had previously been married to Uta Hagen (1938-1948)... \\*
GLIM & He was born in Cork, Ireland, the son of George H. W. Bush and Mary Lynne Morris. \\*
\textbf{Ours} & Despite her poverty, Grace did not marry Whitney Corri Harris, an Irish Catholic... \\
\midrule

\textbf{GT} & In 1962, Clampett created an animated version of the show called Beany and Cecil... \\*
GLIM & Groucho Marx was a notorious snout killer in the Victorian era. \\*
\textbf{Ours} & Libby Ward, the husband of Warren Buffett, has been married to Goldie Hawn... \\
\midrule

\textbf{GT} & When Baldwin was young, he had a job as a busboy at famous New York City disco Studio 54. \\*
GLIM & He was the son of Richard Bush and Mary Fairbanks. \\*
\textbf{Ours} & During the year 1839, he was appointed to the board of the Legal Services Corporation. \\
\midrule

\textbf{GT} & Reagan later married actress Jane Wyman in 1940. \\*
GLIM & He married Laura Welch, a librarian, in 1927. \\*
\textbf{Ours} & During this period, he married Elizabeth Harris, an American actress, on August 10, 2002. \\
\midrule

\textbf{GT} & Friedrich Wilhelm Bessel (July 22, 1784 - March 17, 1846) was a German mathematician... \\*
GLIM & Hepburn was re-elected in 1960 but lost to Republican John Grisham. \\*
\textbf{Ours} & Luca Guido Navier, an American rock musician, was born on May 16, 1965... \\
\midrule

\textbf{GT} & She is married to cinematographer Christopher Duddy (married December 31, 1996 - present). \\*
GLIM & He was born in Los Angeles to Italian immigrants. \\*
\textbf{Ours} & During the following year, she married Michael Jackson, whom she married in 1938, and they later divorced. \\
\midrule

\textbf{GT} & The movie is a blast of educational energy, as bouncy animation and catchy songs... \\*
GLIM & It's a slam dunk that's so bad it's so good it's so good... \\*
\textbf{Ours} & Compared to the earlier, better-looking, and more interesting, Love This Who? is a fun... \\
\midrule

\textbf{GT} & Warm Water Under a Red Bridge is a quirky and poignant Japanese film that explores... \\*
GLIM & The movie is a richly imagined and admirably mature work from a gifted director. \\*
\textbf{Ours} & Leaving the movie a little bit shaky, but it's still a great story, with a great cast and a great story. \\
\midrule

\textbf{GT} & The quirky drama touches the heart and the funnybone thanks to the energetic... \\*
GLIM & This movie is more romantic, emotional, and satisfying than the original. \\*
\textbf{Ours} & Despite its high-quality production, the movie is a little too funny and overly sweet... \\
\midrule

\textbf{GT} & Intriguing and downright intoxicating. \\*
GLIM & The movie is a dull and uninteresting experience. \\*
\textbf{Ours} & ... ... a solid, engaging and engaging film. \\
\midrule

\textbf{GT} & Bad and baffling from the get-go. \\*
GLIM & The movie is too long and overly complicated. \\*
\textbf{Ours} & Despite the lack of a traditional Italian accent, the movie lacks a solid story... \\
\midrule

\textbf{GT} & His nephew is the actor Jack Davenport. \\*
GLIM & He was an American jazz musician. \\*
\textbf{Ours} & an Italian actor played Cyrano in the French movie The Plot. \\
\midrule

\textbf{GT} & A gratingly unfunny groaner littered with zero-dimensional, unlikable characters... \\*
GLIM & It's not a bad movie, but it's not a great one either. \\*
\textbf{Ours} & Despite a grand ambition, the movie is a poor one, with a poor score and a poor acting. \\
\midrule

\textbf{GT} & ... a haunting vision, with images that seem more like disturbing hallucinations. \\*
GLIM & This movie is terrible and boring. \\*
\textbf{Ours} & Despite its young, unimpressive characters, this movie feels like a poorly made satire... \\
\midrule

\textbf{GT} & ... pitiful, slapdash disaster. \\*
GLIM & The movie is genuinely moving. \\*
\textbf{Ours} & The movie looks bad and behaves badly in this bad movie. \\
\midrule

\textbf{GT} & Another Best of the Year selection. \\*
GLIM & It's a film that's a lot of fun. \\*
\textbf{Ours} & ... a genuinely uplifting film. \\

\end{longtable}
\end{small}

%% file: checklist.tex
\section*{NeurIPS Paper Checklist}


\begin{enumerate}

\item {\bf Claims}
    \item[] Question: Do the main claims made in the abstract and introduction accurately reflect the paper's contributions and scope?
    \item[] Answer: \answerYes{} 
    \item[] Justification: Main contributions and scope have been accurately claimed in the abstract and introduction.

\item {\bf Limitations}
    \item[] Question: Does the paper discuss the limitations of the work performed by the authors?
    \item[] Answer: \answerYes{} 
    \item[] Justification: See limitation in Appendix~\ref{app:limitation}.

\item {\bf Theory assumptions and proofs}
    \item[] Question: For each theoretical result, does the paper provide the full set of assumptions and a complete (and correct) proof?
    \item[] Answer: \answerNA{} 
    \item[] Justification: The proposed method does not involve theory assumptions and proof.

    \item {\bf Experimental result reproducibility}
    \item[] Question: Does the paper fully disclose all the information needed to reproduce the main experimental results of the paper to the extent that it affects the main claims and/or conclusions of the paper (regardless of whether the code and data are provided or not)?
    \item[] Answer: \answerYes{} 
    \item[] Justification: We have disclosed all the information, including details of modules, parameters,
implementation details. Our code in attached in supplementary materials.

\item {\bf Open access to data and code}
    \item[] Question: Does the paper provide open access to the data and code, with sufficient instructions to faithfully reproduce the main experimental results, as described in supplemental material?
    \item[] Answer: \answerYes{} 
    \item[] Justification: Our code in attached in supplementary materials with detailed instructions and documentations.

\item {\bf Experimental setting/details}
    \item[] Question: Does the paper specify all the training and test details (e.g., data splits, hyperparameters, how they were chosen, type of optimizer) necessary to understand the results?
    \item[] Answer: \answerYes{} 
    \item[] Justification: See in Appendix~\ref{app:implementation_details}.

\item {\bf Experiment statistical significance}
    \item[] Question: Does the paper report error bars suitably and correctly defined or other appropriate information about the statistical significance of the experiments?
    \item[] Answer: \answerYes{} 
    \item[] Justification: Discussed in Appendix~\ref{app:seed}

\item {\bf Experiments compute resources}
    \item[] Question: For each experiment, does the paper provide sufficient information on the computer resources (type of compute workers, memory, time of execution) needed to reproduce the experiments?
    \item[] Answer: \answerYes{} 
    \item[] Justification: See in Appendix~\ref{app:training}Our code in attached in supplementary
    
\item {\bf Code of ethics}
    \item[] Question: Does the research conducted in the paper conform, in every respect, with the NeurIPS Code of Ethics \url{https://neurips.cc/public/EthicsGuidelines}?
    \item[] Answer: \answerYes{} 
    \item[] Justification: The paper adheres to the NeurIPS Code of Ethics.

\item {\bf Broader impacts}
    \item[] Question: Does the paper discuss both potential positive societal impacts and negative societal impacts of the work performed?
    \item[] Answer: \answerYes{} 
    \item[] Justification: Included in Appendix~\ref{app:limitation}
    
\item {\bf Safeguards}
    \item[] Question: Does the paper describe safeguards that have been put in place for responsible release of data or models that have a high risk for misuse (e.g., pre-trained language models, image generators, or scraped datasets)?
    \item[] Answer: \answerNA{} 
    \item[] Justification: There are no released models and scraped datasets.

\item {\bf Licenses for existing assets}
    \item[] Question: Are the creators or original owners of assets (e.g., code, data, models), used in the paper, properly credited and are the license and terms of use explicitly mentioned and properly respected?
    \item[] Answer: \answerYes{} 
    \item[] Justification: We have properly mentioned the used models and cited them without violating
their license.

\item {\bf New assets}
    \item[] Question: Are new assets introduced in the paper well documented and is the documentation provided alongside the assets?
    \item[] Answer: \answerYes{} 
    \item[] Justification: The code is with the MIT license.

\item {\bf Crowdsourcing and research with human subjects}
    \item[] Question: For crowdsourcing experiments and research with human subjects, does the paper include the full text of instructions given to participants and screenshots, if applicable, as well as details about compensation (if any)? 
    \item[] Answer: \answerNA{} 
    \item[] Justification: The paper does not involve crowdsourcing nor research with human subjects.

\item {\bf Institutional review board (IRB) approvals or equivalent for research with human subjects}
    \item[] Question: Does the paper describe potential risks incurred by study participants, whether such risks were disclosed to the subjects, and whether Institutional Review Board (IRB) approvals (or an equivalent approval/review based on the requirements of your country or institution) were obtained?
    \item[] Answer: \answerNA{} 
    \item[] Justification: The paper does not involve crowdsourcing nor research with human subjects.

\item {\bf Declaration of LLM usage}
    \item[] Question: Does the paper describe the usage of LLMs if it is an important, original, or non-standard component of the core methods in this research? Note that if the LLM is used only for writing, editing, or formatting purposes and does \emph{not} impact the core methodology, scientific rigor, or originality of the research, declaration is not required.
    \item[] Answer: \answerNA{} 
    \item[] Justification:The core method development in this research does not involve LLMs as any
important, original, or non-standard components.

\end{enumerate}